\newif\ifdoubleblind
\newif\iflackofspace

\documentclass[conference]{IEEEtran}
\DeclareRobustCommand{\IEEEauthorrefmark}[1]{\smash{\textsuperscript{\footnotesize #1}}}
\usepackage{cite}

\usepackage{pifont} %
\newcommand{\cmark}{\ding{51}}%
\newcommand{\xmark}{\ding{55}}%

\usepackage{multirow}
\usepackage{tabularx}   %

\usepackage[hidelinks]{hyperref}
\usepackage{flushend} %

\usepackage{orcidlink} %

\usepackage[font=footnotesize,labelfont=footnotesize]{caption}
\usepackage{enumitem}
\setlist[itemize]{noitemsep, topsep=0pt}
\usepackage{booktabs}
\let\originalbottomrule\bottomrule
\renewcommand{\bottomrule}{\addlinespace[0pt]\originalbottomrule}
\let\originalmidrule\midrule
\renewcommand{\midrule}{\addlinespace[0pt]\originalmidrule}
\usepackage{algorithm}
\usepackage{algpseudocode}

\usepackage{amsmath}
\usepackage[capitalise,noabbrev]{cleveref}

\usepackage{float}
\usepackage[caption=false]{subfig}
\usepackage{listings}   %
\newfloat{lstfloat}{htbp}{lop}
\floatname{lstfloat}{Listing}
\crefalias{lstfloat}{listing}

\usepackage{cprotect}
\usepackage{listings}

\usepackage[most]{tcolorbox}
\usepackage{xcolor}

\definecolor{lightlightgray}{rgb}{0.9,0.9,0.9}
\definecolor{codegreen}{rgb}{0,0.6,0}
\definecolor{codegray}{rgb}{0.5,0.5,0.5}
\definecolor{codepurple}{rgb}{0.58,0,0.82}
\definecolor{codeblue}{rgb}{0.04,0.19,0.41}
\definecolor{codered}{rgb}{0.81,0.13,0.18}
\definecolor{backcolour}{rgb}{0.95,0.95,0.95}

\lstdefinestyle{cudastyle}{
commentstyle=\color{codegreen},
keywordstyle=\color{codered},
numberstyle=\tiny\color{codegray},
stringstyle=\color{codeblue}, %
language=C++,                   %
numbers=left,                 %
numberstyle=\scriptsize,      %
stepnumber=1,                 %
numbersep=5pt,                %
showspaces=false,             %
showstringspaces=false,       %
showtabs=false,               %
otherkeywords={__global__,__shared__,__constant__}, %
morekeywords={half,half2},
frame=none,                   %
framerule=0pt,
framesep=0pt, %
xleftmargin=7pt,       %
tabsize=2,                    %
captionpos=b,                 %
breaklines=true,              %
breakatwhitespace=false,      %
escapeinside={\|}{\|},        %
linewidth=1.0\linewidth,     %
basicstyle=\scriptsize\ttfamily,
}

\newtcblisting{cuda}[1][]{%
top=-1mm,
bottom=-1mm,
left=1mm,
right=1mm,
colframe=white,
listing options={
    style=cudastyle,
    belowskip=0pt,
},
#1}

\lstdefinestyle{pythonstyle}{
backgroundcolor=\color{backcolour},
commentstyle=\color{codegreen},
keywordstyle=\color{codered},
numberstyle=\tiny\color{codegray},
stringstyle=\color{codeblue}, %
language=Python,              %
basicstyle=\scriptsize,       %
numbers=left,                 %
numberstyle=\scriptsize,      %
stepnumber=1,                 %
showspaces=false,             %
showstringspaces=false,       %
showtabs=false,               %
frame=none,                   %
frame=none,                   %
framerule=0pt,
framesep=0pt, %
breaklines=true,
tabsize=4,                    %
captionpos=b,                 %
escapeinside={\|}{\|},        %
linewidth=1.03\linewidth,       %
basicstyle=\fontsize{5.4}{2}\ttfamily, %
}%

\newtcblisting{python}[1][]{%
top=-1mm,
bottom=-1mm,
left=1mm,
right=1mm,
colframe=white,
listing options={
    style=pythonstyle,
},
#1}

\lstdefinestyle{snippetstyle}{
commentstyle=\color{codegreen},
keywordstyle=\color{codered},
numberstyle=\tiny\color{codegray},
stringstyle=\color{codeblue}, %
language=C++,                   %
showspaces=false,             %
showstringspaces=false,       %
showtabs=false,               %
otherkeywords={__global__,__shared__,__constant__}, %
morekeywords={half,half2},
frame=none,                   %
framerule=0pt,
framesep=0pt, %
tabsize=4,                    %
captionpos=b,                 %
breaklines=true,              %
breakatwhitespace=false,      %
escapeinside={\|}{\|},        %
linewidth=\linewidth,     %
basicstyle=\footnotesize\ttfamily,
}

\newtcblisting{snippet}[1][]{%
top=-1mm,
bottom=-1mm,
left=1mm,
right=1mm,
colframe=white,
listing options={
    style=snippetstyle,
    belowskip=0pt,
},
#1}

\begin{document}

\title{Constraint-aware Optimization in Auto-Tuning}

\ifdoubleblind
    \author{\IEEEauthorblockN{Anonymous Authors}}
\else

\author{
    \IEEEauthorblockN{Floris-Jan Willemsen\IEEEauthorrefmark{1}\orcidlink{0000-0003-2295-8263}, Stijn Heldens\IEEEauthorrefmark{2}\orcidlink{0000-0001-8792-6305}, Rob V. van Nieuwpoort\IEEEauthorrefmark{1}\orcidlink{0000-0002-2947-9444}, 
Ben van Werkhoven\IEEEauthorrefmark{1}\orcidlink{0000-0002-7508-3272}}
    \IEEEauthorblockA{\IEEEauthorrefmark{1}LIACS, Leiden University, the Netherlands
    \{f.q.willemsen, r.v.van.nieuwpoort, b.van.werkhoven\}@liacs.leidenuniv.nl}
    \IEEEauthorblockA{\IEEEauthorrefmark{2}Netherlands eScience Center, Amsterdam, the Netherlands, s.heldens@esciencecenter.nl}
}

\fi

\maketitle

\thispagestyle{plain}
\pagestyle{plain}

\begin{abstract}
Automatic performance tuning, or auto-tuning, is a key technique in high-performance computing, enabling applications to adapt to complex and evolving hardware architectures. 
A central challenge is the need to optimize over large discrete, constrained parameter spaces, where many candidate configurations are invalid due to hardware or software correctness constraints. 
Traditional evolutionary algorithms, such as Differential Evolution, Particle Swarm Optimization, and Genetic Algorithms, are not inherently constraint-aware and thus often waste computational resources evaluating invalid solutions.

In this work, we present and evaluate constraint-aware variants of four evolutionary algorithms for auto-tuning. 
Through extensive experiments on a representative benchmark suite, we show that constraint-aware optimization leads to faster convergence and improved performance over unconstrained methods. 
Furthermore, we demonstrate that our methods outperform the pyATF methods, a state-of-the-art framework for constraint-based auto-tuning. 
Our results demonstrate that incorporating constraint-awareness into the optimization process significantly enhances their applicability and effectiveness in real-world auto-tuning problems. 
Constraint-awareness improved algorithm efficiency by ${\sim}39\%$ on average, correlated with search space sparsity. 
The algorithms developed in this study are publicly available as open-source contributions to the Kernel Tuner framework, facilitating future research and benefitting users.
\end{abstract}

\IEEEpeerreviewmaketitle

\section{Introduction}

\IEEEPARstart{A}{utomatic} performance tuning, or auto-tuning, is a critical technique in high-performance computing (HPC), enabling developers to optimize software efficiently for specific hardware and input configurations~\cite{balaprakash2017autotuning,vanwerkhoven2020lessons}.
Generic auto-tuning frameworks have been created to provide an application-independent approach for users to create tunable applications, in which performance-critical parameters, or {\em tunable parameters}, such as the number of threads, work per thread, and data layouts, can be varied~\cite{OpenTuner, CLTune, ATF, vanwerkhovenKernelTunerSearchoptimizing2019}.
Auto-tuners systematically explore the vast discrete search space of code variants, generated using metaprogramming or compilation techniques, to automatically identify the optimal parameter configurations (or \textit{solutions}) that maximize performance~\cite{hijma2023optimization, FFTW1998, atlas2001}, energy efficiency~\cite{schoonhovenGoingGreenOptimizing2022}, or other relevant metrics.

A key problem in auto-tuning is the fact that not all code variants constitute feasible (or valid) implementations. 
Modern massively parallel architectures are highly complex with deep memory hierarchies and heterogeneous compute cores, which introduce dependencies between different tunable parameters in the code. 
For example, when applying loop blocking, the tile size of the outer loop has to be a multiple of the tile size used in the inner loop.
Hence, Auto-tuning frameworks allow users to specify {\em constraints} along with the tunable parameters of their applications.
Such constraints significantly complicate the exploration process and impose additional challenges on optimization algorithms~\cite{ATF, BaCO2024}, in which a key problem is that many variants violating the constraints cannot be evaluated due to hardware limitations or software correctness requirements.

Existing optimization algorithms that do not explicitly handle constraints may waste significant computational resources exploring invalid or suboptimal configurations, which can greatly reduce overall performance and efficiency.
Effectively handling constraints within auto-tuning is therefore crucial, yet remains challenging due to the inherent blindness of classical optimization methods, such as evolutionary algorithms (\textit{EAs}) and other metaheuristics, to the feasibility of generated solutions. Traditional optimization operators typically produce candidate solutions irrespective of constraint satisfaction, necessitating specialized constraint-handling techniques such as penalty methods, constraint-specific operators, or repair mechanisms~\cite{Coello2002}. While constrained optimization techniques have clear benefits, their impact on the performance of optimization algorithms for auto-tuning has, to the best of our knowledge, not yet been studied.

To this end, our paper addresses the critical challenge of efficiently incorporating constraint-awareness into auto-tuning optimization algorithms and studying the impact of these techniques on optimization algorithm performance. Specifically, we investigate the integration of constraint-handling capabilities into four widely-used evolutionary optimization algorithms (Differential Evolution, Particle Swarm Optimization, Firefly, and Genetic Algorithm), to systematically avoid or repair invalid solutions during the search, and thereby enhance their performance when solving constrained optimization problems encountered in auto-tuning in particular. 

This paper presents an application study of auto-tuning for performance optimization, taking real-world constraints into account. 
In particular, we make the following contributions:
\begin{itemize}
  \item We review the uptake of techniques developed for constrained optimization in state-of-the-art auto-tuning frameworks.
  \item We present four evolutionary constraint-aware optimization algorithms designed for automatic performance tuning as part of a generic auto-tuning framework.
  \item We quantify the performance impact of using constraint-aware optimization algorithms over traditional optimization algorithms for a representative benchmark~\cite{torring2023towards} set of auto-tuning problems, demonstrating substantial performance improvements across various tuning scenarios. %
  \item We present a performance comparison of our methods with pyATF~\cite{pyATF}, a recently published state-of-the-art framework for constraint-based auto-tuning. 
  \item We have implemented our methods as open-source contributions to Kernel Tuner, a widely-used open-source auto-tuning framework with a broad range of features.
\end{itemize}

The remainder of this paper is organized as follows.
\Cref{sec:background} provides background on the need for constrained optimization in the auto-tuning domain. 
\Cref{sec:related_work} reviews related work regarding constrained optimization in auto-tuning. 
\Cref{sec:design} presents the implementation of our constraint-aware optimization algorithms for auto-tuning. 
\Cref{sec:evaluation} evaluates the impact of constraint-awareness on the performance of optimization algorithms in auto-tuning. 
\Cref{sec:conclusion} concludes. 

\section{Background and Motivation}\label{sec:background}

\begin{lstfloat}[b]
\begin{cuda}[listing only]
__global__ void gemm(const float *A, const float *B, float *C, int M, int N, int K, float alpha, float beta) {
  int row = blockIdx.y * blockDim.y + threadIdx.y;
  int col = blockIdx.x * blockDim.x + threadIdx.x;

  if (row < M && col < N) { |\label{line:gemm-bounds-check}|
    float sum = 0.0f;
    for (int e = 0; e < K; e++) {
      sum += A[row * K + e] * B[e * N + col];
    }
    C[row * N + col] = alpha * sum + beta * C[row * N + col];
  }
}
\end{cuda}
\caption{Example GEMM kernel in HIP/CUDA.}
\vspace{-0.3cm}
\label{code:gemm-naive}
\end{lstfloat}

This section provides a general introduction to auto-tuning using an example kernel to illustrate how constraints arise when creating tunable applications for modern, highly parallel architectures (such as, but not limited to, GPUs).
In this paper, we focus on compile-time auto-tuning where the application can be tuned as part of the development process, as opposed to run-time auto-tuning where the application is tuned while it is running in production~\cite{KTTSoftwareX}.

We will use a simplified general dense matrix-matrix multiplication (GEMM) as our example kernel. GEMM is part of the BLAS linear algebra 
specification, and is a fundamental and widely-used routine in high-performance computing and AI workloads. 
GEMM implements the multiplication of two matrices, $A$ and $B$:
\begin{equation}\nonumber
C = \alpha A \cdot B + \beta C
\end{equation}
where $\alpha$ and $\beta$ are scalars and $C$ is the output matrix. 
$A$ is of size $M\times K$, $B$ is of size $K\times N$, and $C$ is of size $M\times N$. 

\Cref{code:gemm-naive} shows a naive implementation of GEMM as a GPU kernel in HIP/CUDA. This kernel can be executed on a massively parallel GPU processor by a large number of threads in parallel. Specifically, a two-dimensional array $(x, y)$ of threads of size $M\times N$ allows each element in $C$ to be computed by one thread.

GPU programming models, such as HIP or CUDA, do not allow the application developer to directly specify the total number of threads. Instead, threads are organized into {\em thread blocks}, and it is required to specify the block dimensions and the total number of blocks, also referred to as the {\em grid dimensions}. This means that the total number of threads may exceed the number of elements in $C$, which is why on Line~\ref{line:gemm-bounds-check} a check is performed to prevent out-of-bounds array access.

For the GEMM kernel in \cref{code:gemm-naive}, the number of threads per block does not matter for the output of the kernel, but these numbers do impact the performance of the kernel. Thus, the thread block dimensions in both $x$ and $y$ are our first {\em tunable parameters} that constitute functionally equivalent variants of our implementation.

However, to ensure sufficient parallelism, each thread block must have a minimum size of, for example, 32 threads. 
In addition, most parallel architectures pose an upper bound on the number of threads per block, which can be queried before tuning. Typically, this limit is 1024 threads.
These restrictions on the two parameters together form our first constraint: \\ {\small \texttt{32 <= thread\_block\_x * thread\_block\_y <= 1024}}. It is important to understand that this is a {\em hard constraint} as any candidate solution that exceeds 1024 threads per block cannot execute. Therefore, the execution time, i.e., the value of our cost function, cannot be obtained for candidate solutions violating this constraint.

The kernel shown in \cref{code:gemm-naive} is a so-called naive kernel. A highly-optimized implementation would require extensive optimization, such as the use of tensor cores and other modifications, each introducing more tunable parameters along with more constraints. For example, to efficiently exploit the cache hierarchy and specialized memory spaces in modern architectures, we would need to introduce {\em loop blocking}. Loop blocking, in turn, introduces more constraints; for example, the hardware limits of specialized memory spaces such as shared memory, and the loop count needs to be a divisor of the total work assigned to a thread block or a higher-level loop-blocking scheme.
As a full review of all applied code transformation techniques and their constraints for a highly-tunable GEMM implementation is beyond the scope of this paper, we refer the reader to T{\o}rring et al.~\cite{torring2023towards} for a more extensive description. 
For a comprehensive overview of code transformation techniques, we refer to Hijma et al.~\cite{hijma2023optimization}.

\section{Related Work}\label{sec:related_work}

\begin{table*}[tb!]
\caption{Overview of support for constrained optimization in related and prior work. SA = Simulated Annealing. PSO = Particle Swarm Optimization.}
\label{tab:related_work}
\centering
\small
\setlength{\tabcolsep}{2pt}
\begin{tabular}{p{2cm}|p{1.5cm}|p{2.1cm}|p{8cm}|p{3.5cm}}
\toprule
\textbf{Tool}	& \textbf{Supports constraints}	& \textbf{Search\,Space Representation}	& \textbf{Optimization algorithms} & \textbf{Constraint handling}	\\
\midrule
AUMA~\cite{AUMA}	& \xmark	& list	& K-Bagging Neural Networks	& Only feasible solutions	\\
\hline
OpenTuner~\cite{OpenTuner}	& \xmark	& list	& SA, PSO, Bandit, various Evolutionary Algorithms, Local Search, Random, Exhaustive	& Left to the user \\
\hline
CLTune~\cite{CLTune}	& \cmark	& list	& SA, PSO, Neural Network, Random, Exhaustive	& Only feasible solutions	\\
\hline
ytopt~\cite{ytopt}	& \cmark	& ConfigSpace	& Bayesian Optimization	& Left to the user	\\
\hline
GPTune~\cite{liuGPTuneMultitaskLearning2021}	& \cmark	& scikit-optimize.space	& Bayesian Optimization	& Single constant penalty \\ %
\hline
KTT~\cite{KTTSoftwareX}	& \cmark	& chain-of-trees	& Markov Chain Monte Carlo, Profile-based search, Random, Exhaustive	& Only feasible solutions	\\
\hline
ATF~\cite{ATF} \newline pyATF~\cite{pyATF}	& \cmark	& chain-of-trees	& SA, Differential Evolution, Bandit, Local Search, Random, Exhaustive	& Only feasible solutions	\\
\hline
BaCO~\cite{BaCO2024}	& \cmark	& chain-of-trees	& Bayesian Optimization, Exhaustive	& Only feasible solutions. Surrogate model for hidden constraints.	\\
\hline
Kernel\,Tuner	& \cmark	& list	& 20 different global and local optimization algorithms, including Annealing methods, Evolutionary / Genetic methods, Swarm-based methods, and Bayesian Optimization	& Single constant penalty or only feasible solutions	\\
\bottomrule
\end{tabular}
\end{table*}

This section provides an overview of the application of constrained optimization in state-of-the-art auto-tuning frameworks.
\cref{tab:related_work} provides an overview of the use of constrained optimization methods in auto-tuning frameworks.
AUMA~\cite{AUMA} does not support constraints directly but relies on an external tool to generate the list of valid configurations. OpenTuner~\cite{OpenTuner} does not support constraints. 

CLTune~\cite{CLTune} maintains the full list of feasible configurations in memory but does support constraints. %
CLTune implements SA, PSO, and a Neural Network-based search algorithm in addition to exhaustive and random search. The search space representation is used to ensure only valid points are evaluated.
For example, when using PSO, the movement of particles is limited to only valid neighboring configurations. The particle movement step is simply repeated until a valid configuration is found. There is also a probability that the particle does not move at all. To check the validity of configurations, CLTune simply iterates through the entire list of valid configurations until it finds the matching configuration, if the configuration is not found it is assumed to be invalid. Similarly, when moving to a neighboring configuration during Simulated Annealing, CLTune iterates through the entire list to find all neighbors of a configuration.
Despite support for constrained optimization, none of the optimization algorithms implemented in CLTune outperformed random search in their evaluation~\cite{CLTune}. 

GPTune and ytopt rely on \verb|scikit-optimize.space| and \verb|ConfigSpace|, respectively, which represent multidimensional configuration spaces, but do not enumerate or store individual configurations. Instead, these provide an interface to generate random samples from the search space, after which the validity of the drawn sample is checked. As constraint resolution is not supported by \verb|scikit-optimize.space|, GPTune relies on an additional internal check on sampled points.
OpenTuner~\cite{OpenTuner} and ytopt are designed as libraries that allow users to implement auto-tuners. As such, the entire implementation of how to compile and benchmark possible solutions, and how to handle the evaluation of invalid configurations, is left to the user. For example, some examples of ytopt show that infinity is returned instead of the execution time when a selected configuration does not compile successfully. 

Most auto-tuning frameworks supporting user-defined constraints check the feasibility of candidate solutions as part of search space construction. ATF~\cite{ATF}, KTT~\cite{KTTSoftwareX}, BaCO~\cite{BaCO2024}, and pyATF~\cite{pyATF} use chain-of-trees to store the set of valid configurations in memory.

ATF~\cite{ATF} introduced the chain-of-trees search space representation and construction method that has since been adopted by many auto-tuning frameworks.
The same authors have recently presented pyATF~\cite{pyATF} as a new state-of-the-art framework for constraint-based auto-tuning.
pyATF and ATF both include some optimization algorithms, all of which operate on a continuous domain $[0, 1]^N$, where $N$ is the number of tunable parameters. The objective function maps the continuous coordinates back to the nearest valid configuration in the chain-of-trees representation. 
However, the construction of the chain-of-trees representation is potentially expensive~\cite{searchspace_construction_paper}.

KTT~\cite{KTTSoftwareX} is a C++ auto-tuning framework with a focus on runtime auto-tuning. KTT also uses chain-of-trees and its optimization algorithms operate directly on the indices into the chain-of-trees. The algorithms are designed to only sample from the feasible solutions when performing random sampling or retrieving the list of neighboring configurations.

BaCO, like ATF, samples only from valid configurations based on the user-defined constraints. However, BaCO takes an interesting approach to also learn so-called {\em hidden constraints}. Some configurations might appear to be valid according to user-defined constraints, but turn out to be infeasible during compilation or at run time, and are thus considered to violate hidden constraints. A separate surrogate model is trained to predict the feasibility of solutions according to the hidden constraints and the acquisition function is modified to take the probability of feasibility into account.

Kernel Tuner is an open-source Python-based auto-tuning framework that implements many different optimization algorithms, including Differential Evolution, Particle Swarm Optimization, and Genetic Algorithm.
The optimization algorithms are responsible for selecting the next code variants to be compiled and benchmarked on the GPU. 
Before the modifications presented in this paper, if a code variant selected by the optimizer was not valid under the constraints, Kernel Tuner applied a simple static penalty, e.g., $-10^{20}$~\cite{schoonhoven2022benchmarking}.
In this work, we enhance this and extend Kernel Tuner with several new constraint-aware optimization algorithms to improve its performance for constraint-based auto-tuning.

\section{Design and Implementation}\label{sec:design}

This section presents an overview of the design and implementation of our methods implemented in Kernel Tuner\footnote{\url{https://github.com/KernelTuner/kernel_tuner}}. 
The overall auto-tuning problem can be formalized as an optimization problem which determines the optimal code variant $v^\star$ (assuming maximization) as follows:
\begin{equation} \label{eq:autotuning}
v^\star = \underset{v\in\mathcal{V}}{\text{arg max}} \, f_{H_j,I_k}(A_{i,v})
\end{equation}
Where we have an application $A_i$ on a hardware platform $H_j$ for an input data set $I_k$ to maximize the performance measured by $f_{H_j,I_k}(A_i)$ over the code variants in a search space $\mathcal{V}$.

\subsection{Kernel Tuner}\label{sec:kernel-tuner}

\begin{figure}
    \centering
    \includegraphics[width=0.95\linewidth]{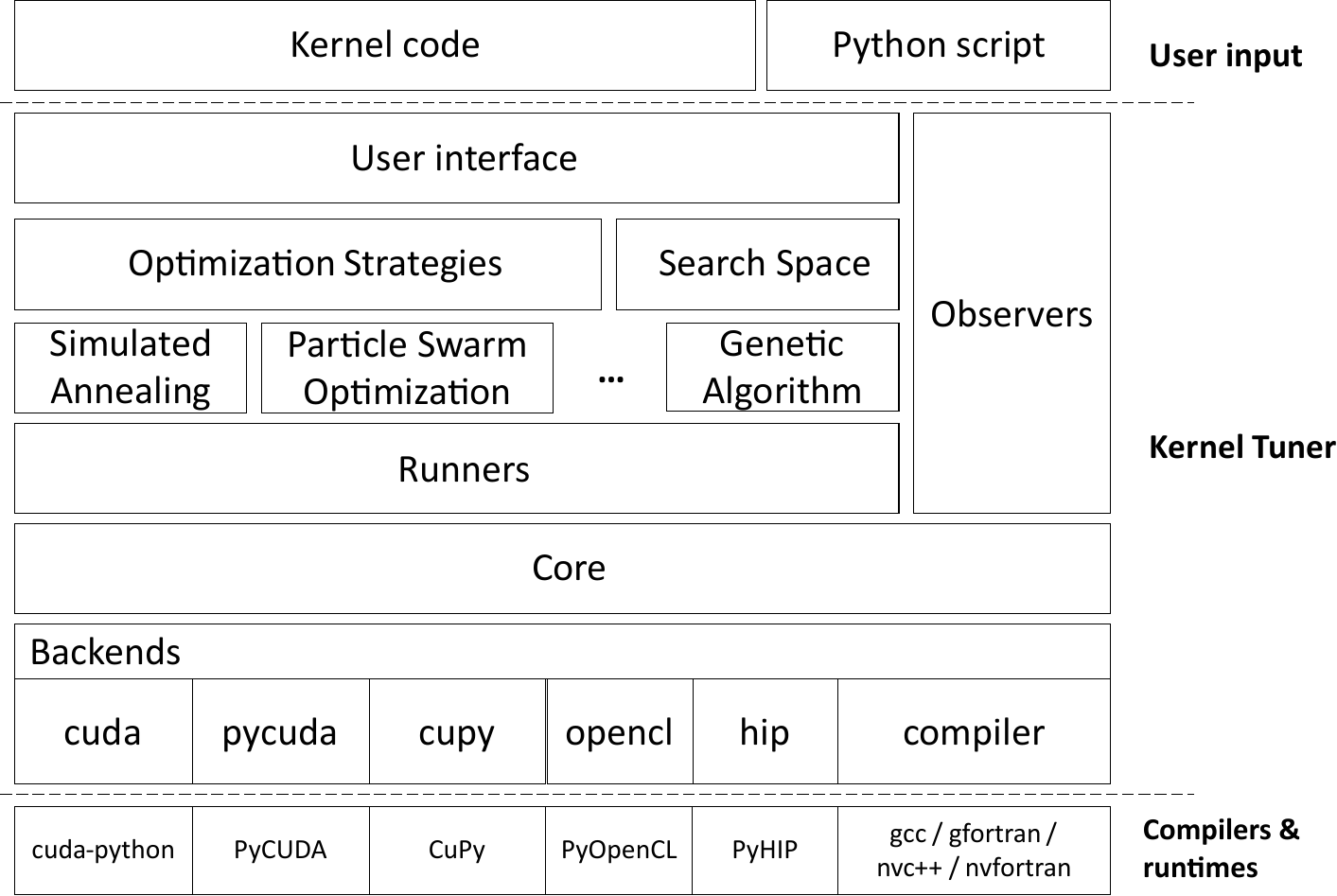}
    \caption{Overview of the software architecture of Kernel Tuner.}
    \label{fig:kt-arch}
    \vspace{-0.3cm}
\end{figure}

Kernel Tuner is generally used as an external framework for developers to benchmark and optimize GPU kernels in isolation, which can then be used with applications in any host programming language.  An overview of the software architecture of Kernel Tuner is shown in \cref{fig:kt-arch}.
Users of Kernel Tuner create a small Python script that points to the kernel code and describes the data set used for benchmarking, the tunable parameters that constitute the code variants, and the constraints. 
There are also various optional settings that users can specify, such as derived metrics to be computed, the optimization objective to use, which optimization algorithm to use, and hyperparameters to this optimization algorithm.

The optimization strategy uses the search space to select new candidate solutions for evaluation. Some optimization algorithms have hyperparameters such as the annealing schedule or the number of generations, which control when to stop the optimization process. However, Kernel Tuner interrupts an optimization algorithm when the user-specified optimization budget is exceeded. The runners are responsible for the actual evaluation of candidate solutions, which means compiling and measuring the performance of the code variants on the GPU. The performance of each candidate solution is measured multiple times, and the average execution time is returned to the optimization algorithm.

The runner uses a single high-level interface inside Kernel Tuner's core layer that interfaces to the low-level backends. The backends, in turn, interface with the Python bindings of the compilers and device runtimes, such as CUDA, OpenCL, and HIP~\cite{lurati2024bringing}. Also shown in \cref{fig:kt-arch} are the Observers which facilitate the observation of metrics besides execution time, such as the energy consumed by the GPU~\cite{powersensor3} or the numerical accuracy of the computation~\cite{accuracytuning}.

In Kernel Tuner, the auto-tuning search space construction problem is formalized as a Constraint Satisfaction Problem (\textit{CSP}) \cite{searchspace_construction_paper} defined by $\mathcal{P} = (X, D, C)$, where:
\begin{itemize}
  \item $X = \{x_1, x_2, \dots, x_n\}$ is a finite set of variables, each corresponding to a tuning parameter (e.g., block size, tile width).
  \item $D = \{D_1, D_2, \dots, D_n\}$ is a set of finite domains, where $D_i$ is the set of legal values for variable $x_i$.
  \item $C = \{c_1, c_2, \dots, c_m\}$ is a finite set of constraints, where each $c_j$ is a predicate over a subset of variables $\text{scope}(c_j) \subseteq X$ that restricts the allowable combinations of values based on hardware limits or algorithmic correctness.
\end{itemize}
A solution to the auto-tuning search space is then a total assignment $ \mathcal{V}: X \rightarrow \bigcup D_i$ such that $\mathcal{V}(x_i) \in D_i$ for all $i$, and all constraints $c_j \in C$ are satisfied under $\mathcal{V}$.
What distinguishes the auto-tuning problem from other constrained optimization problems is that $C$ generally contains {\em hard constraints}, meaning that the performance function $f_{H_j,I_k}(A_{i,v})$ cannot be evaluated for any $v$ that does not satisfy the constraints.

Kernel Tuner determines all configurations in the search space $\mathcal{V}$ before starting the tuning process.
This is helpful as important search space characteristics, such as the true parameter bounds, can guide optimization algorithms more effectively and facilitate the use of stratified sampling techniques, such as Latin Hypercube Sampling~\cite{willemsenBayesianOptimizationAutotuning2021}. 
In addition, the full search space resolution substantially reduces the cost of valid neighbor lookups, which is important for several optimization algorithms that use these operations extensively. 
Thanks to the use of an optimized constraint satisfaction problem solver, this step can be performed with minimal impact on the total execution time~\cite{searchspace_construction_paper}.

The \textit{Search Space} object in Kernel Tuner provides a single interface for all search space-related operations that can be used by optimization algorithms to navigate the search space in a constraint-aware manner. To this end, \textit{Search Space} contains functionality to query whether a particular solution is valid, to generate a number of random samples of only valid solutions, and to get all valid neighboring solutions of a particular solution.
There are different ways to define neighbors.
Currently, we have implemented four definitions that specify when two solutions qualify as \emph{neighbors}:

\begin{itemize}
    \item \textbf{Strictly adjacent:} For every parameter, their values are identical or the previous/next value.
    \item \textbf{Adjacent:} For each parameter, their values are either identical or the nearest previous/next value that gives a valid configuration.
    \item \textbf{Hamming:} All parameters are identical except one.
    \item \textbf{Index-distance}: Minimizes the sum of absolute differences between parameter value indices, returning all configurations with the lowest distance as neighbors. 
\end{itemize}

The following subsections describe the implementations of our constraint-aware optimization algorithms in Kernel Tuner. Specifically, we focus on how the algorithms use the search space to optimize the auto-tuning problem in the presence of hard constraints.

\subsection{Differential Evolution}

\begin{algorithm}[tb]
\footnotesize
\caption{Differential Evolution}
\label{alg:differential-evolution}
\begin{algorithmic}[1]
\Require Population $X$, Space $\mathcal{V}$, Scaling $F \in [0, 2]$, Crossover $CR \in [0, 1]$

\State $X \gets \{x_1, x_2, ..., x_N\} \in \mathcal{V}$ \Comment{Initial population}
\For{$x_i \in X$} \Comment{Main evolution loop}
    \State $v_i \gets$ \textsc{Mutate}($x_i, X, best$)
    \State $u_i \gets$ \textsc{Crossover}($x_i, v_i, CR$)
    \State $u_i \gets$ \textsc{Repair}($u_i, \mathcal{V}$)
    \If{$f(u_i) \leq f(x_i) \text{ and } u_i \notin X$}
        \State $x_i \gets u_i$ \Comment{Selection}
        \If{$f(u_i) \leq best$} 
            \State $best \gets u_i$ \Comment{Update best}
        \EndIf
    \EndIf
\EndFor

\Procedure{Mutate}{$x_i, X, best$}
    \State $\{x_{r1}, x_{r2}, x_{r3}\} \gets \text{select 3 random vectors from } X \setminus \{x_i\}$
    \If{use\_best}
        \State $x_{r1} \gets best$
    \EndIf
    \State $\{i_{r1},i_{r2},i_{r3}\} \gets to\_indices(\{x_{r1}, x_{r2}, x_{r3}\})$
    \State $m \gets i_{r1} + F \cdot (i_{r2} - i_{r3})$ \Comment{Apply mutation}
    \State \Return $to\_values(round\_and\_clip(m))$ 
\EndProcedure

\Procedure{Crossover}{$x_i, v_i, CR$}
    \State $j_{rand} \gets \text{random index } \in \{1, \dots, N\}$
    \For{$j \gets 1$ to $N$}
        \State $r \gets \text{random } \in [0, 1]$
        \State $u_{i,j} \gets (r < CR \text{ or } j = j_{rand}) ? v_{i,j} : x_{i,j}$
    \EndFor
    \State \Return $u_i$
\EndProcedure

\Procedure{Repair}{$u_i, \mathcal{V}$}
    \If{$u_i \notin \mathcal{V}$}
        \State $u_i \gets get\_nearest\_neighbor(u_i, \mathcal{V})$
    \EndIf
    \State \Return $u_i$
\EndProcedure

\end{algorithmic}
\end{algorithm}

The differential evolution strategy in Kernel Tuner is quite flexible and supports many different mutation and crossover operators. \Cref{alg:differential-evolution} shows a simplified pseudocode of the differential evolution strategy using mutation with three candidates, either three random or two random and the current best, and binomial crossover, known as the {\em rand1bin} or {\em best1bin} variants.

The algorithm starts with generating an initial population of valid candidates using Latin Hypercube Sampling (LHS) and the earlier constructed search space $\mathcal{V}$. The four main steps are mutation, crossover, repair and selection. Mutation and crossover generate a set of trial vectors $\{u_1, .., u_N\}$, which are evaluated and replace the corresponding population member $\{x_1, .., x_N\}$ if $u_i$ outperforms $x_i$. Note that $u_i$ only replaces $x_i$ if $u_i$ is not already in $X$. Also, if the population does not change at all over two consecutive generations, the population is reinitialized randomly.

The evaluation of $f(u_i)$ includes compiling and benchmarking the code variant $u_i$ on the GPU. Kernel Tuner uses a memoization scheme to avoid evaluations of candidate solutions that have already been compiled and benchmarked.

Both mutation and crossover run the risk of generating a trial vector that violates the hard constraints on the search space. As such, the repair method is applied to each trial vector after these two steps to change the invalid trial vectors to their nearest valid neighbor in the search space, using the sum of absolute differences between parameter value indices. 

The differential evolution strategy in Kernel Tuner supports all commonly-used mutation operators within differential evolution, including mutation with up to five random candidates, {\em currenttobest} and {\em randtobest} variants, which can all be combined with either binomial or exponential crossover. The non-constraint-aware version generates the initial population randomly and skips the repair step.

\begin{algorithm}[tb]
\footnotesize
\caption{Constraint-aware Repair Method for PSO}
\label{alg:pso-repair}
\begin{algorithmic}[1]
\Require Current position $x \in [0,1]^N$, Search Space $\mathcal{V}$

\Procedure{Neighbors}{$p$}  %
\For{$m \in$ \{`strictly-adjacent', `adjacent', `Hamming'\}}
    \State $n \gets list\_neighbors(p, m) \in \mathcal{V}$
    \If{$|n| > 0$}
        \State \textbf{return} $n$
    \EndIf
\EndFor
\State \textbf{return} []
\EndProcedure

\State $p \gets get\_params(x)$ \Comment{Convert from continuous space}
\State $n \gets$ \textsc{Neighbors}($p$) $\in \mathcal{V}$
\If{$|n| > 0$}
    \State $n \gets to\_coordinates(n)$ \Comment{Convert to continuous space}
    \State $s \gets Euclidian\_distance(x, n)$
    \State $n \gets sort(n, key=s)$ \Comment{Sort neighbors on distance to $x$}
    \State \textbf{return} $n[0]$
\EndIf
\State \Return $[]$
\end{algorithmic}
\end{algorithm}

\subsection{Particle Swarm Optimization (PSO)}

The PSO strategy in Kernel Tuner is a simple and representative implementation of the PSO methods. Starting from a randomly generated sample population of solutions, the inertia, cognitive, and social coefficients are used to move the particles around according to their own best solution and the global best solution found so far. In contrast to Differential Evolution, the particles in PSO are represented and move around in a continuous space. 

To allow continuous optimization algorithms, such as Particle Swarm Optimization, to work on the discrete optimization problem of auto-tuning, discrete points are mapped linearly into a continuous domain $[0, 1]^N$, where $N$ is the number of tunable parameters. First, the values of the tunable parameter with the largest set of possible values are linearly distributed across equidistant points in the domain $[0, 1]$ with distance $\epsilon$. In the other dimensions, values are mapped to points distributed between $[0, m\times \epsilon]$, where $m$ is the number of values in that dimension. This method ensures that, regardless of the number of values in a particular dimension, a perturbation by $\epsilon$ in any dimension leads to a position that represents a different solution in the discrete space.
The continuous coordinates are converted back to discrete points by snapping to the nearest discrete point. 

The position of the particle is not adjusted when converting between continuous and discrete representations. To make PSO efficient for constraint-based auto-tuning, we (i) use the search space to generate a population of only valid candidate solutions, which are then mapped to continuous coordinates as starting positions, and (ii) ensure that when the inverse conversion is needed during evaluation, the continuous coordinates are mapped back to only valid discrete configurations, using the procedure outlined in \cref{alg:pso-repair}. 
When the nearest discrete point is not feasible, we use the search space to retrieve a list of valid neighbors. We first try the \textit{strictly-adjacent}, then \textit{adjacent}, and then \textit{Hamming} neighbor rules. All configurations in the first nonempty set that is returned are converted to continuous space, and the closest point, by Euclidean distance in continuous space, is selected and used in the evaluation of the particle's position. Note that this repair method only affects evaluation and does not change the particle's position itself.

We also implement the Firefly Algorithm as a variant of PSO, using the same sampling and repair techniques to ensure only valid candidate solutions are evaluated. The non-constraint-aware version generates the initial coordinates in $[0,1]^N$ randomly and applies a single static penalty to invalid positions instead of evaluating the closest valid configuration.

\subsection{Genetic Algorithm}

\begin{figure}[t]
    \centering
    \vspace{-0.4cm}
    \includegraphics[width=\linewidth]{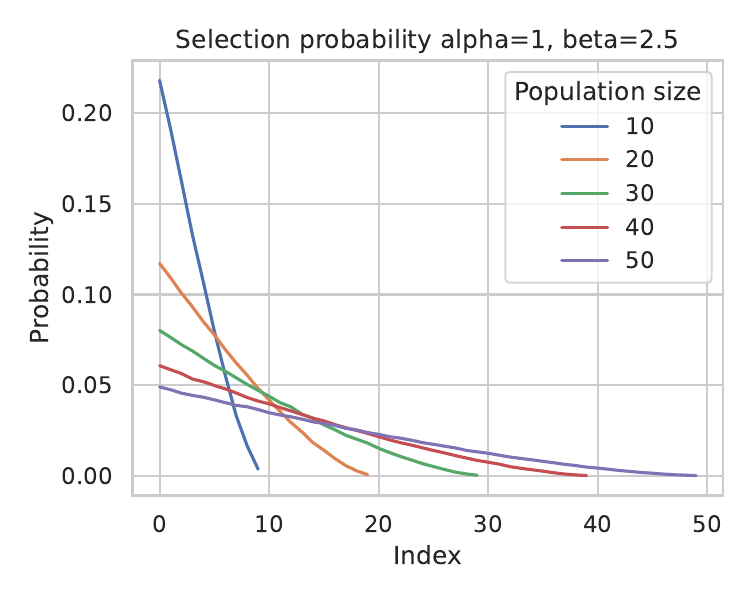}
    \vspace{-0.8cm}
    \caption{Beta-distributed relation between the index in the sorted population and the probability of being selected for crossover for different population sizes.}
    \label{fig:beta-distribution}
\end{figure}

The Genetic Algorithm strategy in Kernel Tuner uses a population of candidate solutions that is completely refreshed every generation. 
The Genetic Algorithm supports single-point, two-point, uniform, and disruptive uniform crossover. Individuals in the population are sorted based on the cost function value. Selection for crossover uses a beta probability distribution to ensure that individuals with better cost function values have a higher probability of being selected, as shown in \cref{fig:beta-distribution}. The advantage of this approach is that selection is biased towards better individuals independently of the magnitude of the objective function values.

After crossover, newly generated individuals may violate the constraints and therefore need to be repaired. \cref{alg:ga-repair} shows the repair procedure used in GA. Reusing the \textsc{Neighbors} procedure from PSO (\cref{alg:pso-repair}), we try the \textit{strictly-adjacent}, then \textit{adjacent}, and then \textit{Hamming} neighbor rules to retrieve the list of neighbors. The repair method replaces the invalid solution with a random neighbor in the first nonempty set that is returned. In contrast to PSO, the repaired configuration replaces the invalid individual in the population. Our GA first repairs invalid solutions and then mutates if needed. 

Mutation happens in exactly one parameter value, where the value is replaced with a different value for that tunable parameter, if any. The constraint-aware GA implements mutation by replacing an individual with a random valid Hamming neighbor.

\begin{algorithm}[tb]
\footnotesize
\caption{Constraint-aware Repair Method for GA}
\label{alg:ga-repair}
\begin{algorithmic}[1]
\Require Current solution $s$, Search Space $\mathcal{V}$
\If{$s \notin \mathcal{V}$}
    \State $n \gets$ \textsc{Neighbors}($s$) $\in \mathcal{V}$
    \If{$|n| > 0$}
        \State \textbf{return} $random\_choice(n)$
    \EndIf
\EndIf
\State \Return $s$
\end{algorithmic}
\end{algorithm}

\section{Evaluation}\label{sec:evaluation}

In this section, we evaluate the effectiveness of the constraint-aware optimization algorithms presented in \cref{sec:design} and quantify their performance improvement. In addition, we present a comparison with pyATF, a state-of-the-art framework for constraint-based auto-tuning~\cite{pyATF}.

\subsection{Experimental setup}\label{subsec:evaluation_experimental_setup}

\begin{table*}[tb]
    \small
    \centering
    \caption{GPUs used in our experiments. *Only one out of two dies of the MI250X is used.}\label{tab:gpu-properties}
    \begin{tabular}{l|l|l|r|r|r|r|r}
    \toprule
    \textbf{GPU} & \textbf{Year} & \textbf{Architecture} & \textbf{Cores} & \textbf{Memory} & \textbf{Cache} & \textbf{Bandwidth (GB/s)} & \textbf{Peak SP (GFLOPS/s)}\\
    \midrule
    AMD W6600    & 2021 & RDNA 2 & 1792 & 16~GB GDDR6 & 32 MB L3 & 224  & 10404 \\
    AMD MI250X*  & 2021 & CDNA 2 & 7040 & 64~GB HMB2e & 8 MB L2  & 1638 & 28160 \\
    AMD W7800  & 2023 & RDNA 3 & 4480 & 32~GB GDDR6 & 64 MB L3  & 576 & 45250 \\
    Nvidia A4000 & 2021 & Ampere & 6144 & 8~GB GDDR6  & 4 MB L2  & 448  & 17800 \\
    Nvidia A6000 & 2020 & Ampere & 10752 & 48~GB GDDR6  & 6 MB L2  & 768  & 38710 \\
    Nvidia A100  & 2020 & Ampere & 6912 & 40~GB HMB2  & 40 MB L2 & 1555 & 19500 \\
    \bottomrule
    \end{tabular}
\end{table*}

For the evaluation, we focus on six different GPU models available in the DAS-6~\cite{DASMediumScaleDistributedSystem} and LUMI~\cite{LUMIsupercomputer} supercomputers.
The GPU specifications are listed in \cref{tab:gpu-properties}. On DAS-6, we use Rocky-8 Linux 4.18, ROCM 6.0.2 with AMD clang 17.0.0, and CUDA 12.2 with GCC 9.4.0. For the MI250X, LUMI is running SUSE Linux 5.14.21, ROCM 5.2.3 with AMD clang 14.0.0. Note that the MI250X is a multi-chip module with two individually operating GPU dies, of which we use only a single die.
The Python version used is 3.11.7. 
All measurements have been performed with Kernel Tuner version 1.3.1 and compared against pyATF version 0.0.9.

\begin{table*}[tb]
    \small
    \centering
    \caption{Overview of the basic characteristics of the real-world applications.}
    \label{tab:searchspaces_real_world_overview}
    \begin{tabularx}{\linewidth}{l|l|l|l|l|X|l}
    \toprule
        \textbf{Name} & \textbf{Cartesian size} & \textbf{Constrained size} & \textbf{Dimensions} & \textbf{No. constraints} &  \textbf{No. values per parameter} & \textbf{Density \%} \\
    \midrule
        Dedispersion    & 22272     & 11130  & 8  & 3 &  1 - 29 & 49.973 \\
        2D Convolution  & 10240     & 4362   & 10 & 4 &  1 - 16  & 42.598 \\ 
        Hotspot         & 22200000  & 349853 & 11 & 5 &  1 - 37 & 1.576\\
        GEMM            & 663552    & 116928 & 17 & 8 &  1 - 4  & 17.622 \\
    \bottomrule
    \end{tabularx}
\end{table*}

We will evaluate our approach using the BAT benchmark suite~\cite{torring2023towards} of auto-tunable GPU kernels. Specifically, we use the \textit{dedispersion}, \textit{convolution}, \textit{hotspot}, and \textit{GEMM} kernels. These benchmark kernels are examples of widely used real-world applications in astronomy, image processing, materials science, and linear algebra, respectively. The characteristics of these applications are shown in~\cref{tab:searchspaces_real_world_overview}. The Cartesian size is the size of the complete combinatorial space without applications of constraints. The constrained size is the number of feasible solutions that remain after applying constraints. The number of dimensions equals the number of tunable parameters in the source code. Finally, the density is the percentage of valid solutions (constrained size over the Cartesian size). We have selected these search spaces to represent a wide range of densities and other characteristics, as we would like to study their influence on optimization algorithm performance for constrained optimization problems.

\textit{Dedispersion} is a signal-processing kernel that reconstructs radio signals distorted by interstellar dispersion by applying a range of dispersion measures to time-domain samples across multiple frequency channels~\cite{sclocco2020amber}.

The \textit{2D convolution} kernel performs image filtering by computing weighted sums over image regions, with tunable parameters for thread block size, work per thread, shared memory padding, and use of the read-only cache.

\textit{Hotspot} is a thermal simulation kernel that estimates processor temperature by iteratively solving differential equations based on simulated power and initial temperature inputs, producing a temperature grid as output. This tunable implementation supports flexible thread/block configurations and temporal tiling.

\textit{GEMM} (General Matrix-Matrix Multiplication) is the example operation that has been introduced in \cref{sec:background}. This tunable GEMM kernel originates from CLBlast~\cite{CLBlast2018}, a tunable linear algebra library. 

These applications also bring diversity in their performance characteristics; e.g., dedispersion and hotspot are generally bandwidth-bound, while convolution and GEMM are generally compute-bound. 
To obtain a diverse set of real-world auto-tuning cases for evaluation, we use these four auto-tuning applications on the six GPUs described in \cref{tab:gpu-properties}, resulting in 24 unique search spaces.

To compare the performance optimization algorithms, we use the methodology for comparing optimization algorithm performance for auto-tuning problems as outlined by the auto-tuning research community~\cite{willemsen2024methodology}. 
This methodology provides a systematic approach to comparing optimization algorithms across auto-tuning search spaces, reflecting the actual total time spent (including resolving the search space) as this is most relevant for real-world usage. Per search space in the comparison set, the approach defines how to set the optimization budget and defines a \emph{calculated performance baseline}, a statistical approximation of \emph{random search} over the feasible solutions.
In particular, it defines a \textit{performance score} $\mathcal{P}$ that quantifies an optimization algorithm's performance over the passed time relative to the calculated baseline, to have consistent, objective-independent, transparent, and comparable behavior across search spaces. 
\begin{equation} \label{eq:performance_score}
\mathcal{P}(\mathcal{F},A,H,I) = \frac{1}{|\mathcal{T}|} \sum_{t \in \mathcal{T}} \frac{\displaystyle \sum_{A_i \in A} \sum_{H_j \in H} \sum_{I_k \in I} \mathcal{P}(\mathcal{F}, A_i, H_j, I_k)_t}{|A||H||I|}
\end{equation}
The applications $A$, target hardware platforms $H$, and inputs $I$ are the collections of $A_i$, $H_j$, and $I_k$ of \cref{eq:autotuning}, $\mathcal{T}$ is the set of sampling points in time used to aggregate performance over time.
This aggregate score enables robust comparison of optimization algorithms by capturing both the quality of the configurations found as well as the time taken to do so. 
As such, a difference when comparing two performance scores can indicate a difference in the quality of configurations found, the time taken to do so, or a combination of both. 
A score of 0.0 indicates that the performance over time is similar to the baseline, whereas a score of 1.0 indicates that the optimum is found immediately. 
For this evaluation section, the allocated budget for each run is equivalent to the time it takes the baseline to reach 95\% of the distance between the search space median and optimum. 
Each optimization algorithm has been run 100 times on each search space to mitigate stochasticity. 

\subsection{Impact of Constrained Optimization for Auto-Tuning} \label{subsec:evaluation_kt}

In this subsection, we investigate the impact of using constrained optimization techniques for auto-tuning applications. To this end, we compare the performance of Differential Evolution (DE), Particle Swarm Optimization (PSO), Firefly Algorithm, and  Genetic Algorithm (GA) with and without the modifications for constrained optimization presented in \cref{sec:design}. %

\begin{table}[tb]
    \centering
    \small
    \caption{Hyperparameter values for the optimization algorithms.}
    \label{tab:hyperparams}
    \begin{tabular}{l|r|r}
        \toprule
        \textbf{Algorithm} & \textbf{Hyperparameter} & \textbf{Values} \\
        \midrule
        \multirow{4}{*}{Differential Evolution (\textit{DE})} & popsize & 16 \\ 
        & differential weight & 0.7 \\ 
        & crossover rate & 0.6 \\ 
        & method & best1bin \\ 
        \midrule
        \multirow{5}{*}{\shortstack{Particle Swarm\\ Optimization (\textit{PSO})}} & popsize & 30 \\ 
        & maxiter & 100 \\ 
        & w & 0.5 \\  
        & c1 & 3.0 \\ 
        & c2 & 0.5 \\
        \midrule
        \multirow{5}{*}{Firefly Algorithm} & popsize & 20 \\ 
        & maxiter & 100 \\ 
        & B0 & 1.0 \\
        & gamma & 1.0 \\
        & alpha & 0.2 \\
        \midrule
        \multirow{4}{*}{Genetic Algorithm (\textit{GA})} & method & single\_point \\ 
        & popsize & 20 \\ 
        & maxiter & 150 \\ 
        & mutation\_chance & 5 \\
        \bottomrule
    \end{tabular}
\end{table}

The hyperparameters of the optimization algorithms are shown in \cref{tab:hyperparams}, with the hyperparameters tuned as per the extended tuning method of \cite{willemsenTuningTheTuner}. Both the constrained and nonconstrained versions of the optimization algorithms use the same hyperparameters. 
The \textit{popsize} parameter in DE is multiplied by the number of dimensions to get the actual population size. 
The mutation chance is interpreted as a `one in X' chance, so a mutation chance of 5 means there is a 0.2 probability of a mutation when generating offspring in the Genetic Algorithm. 

\begin{figure}[tb]
    \centering
    \includegraphics[width=\linewidth]{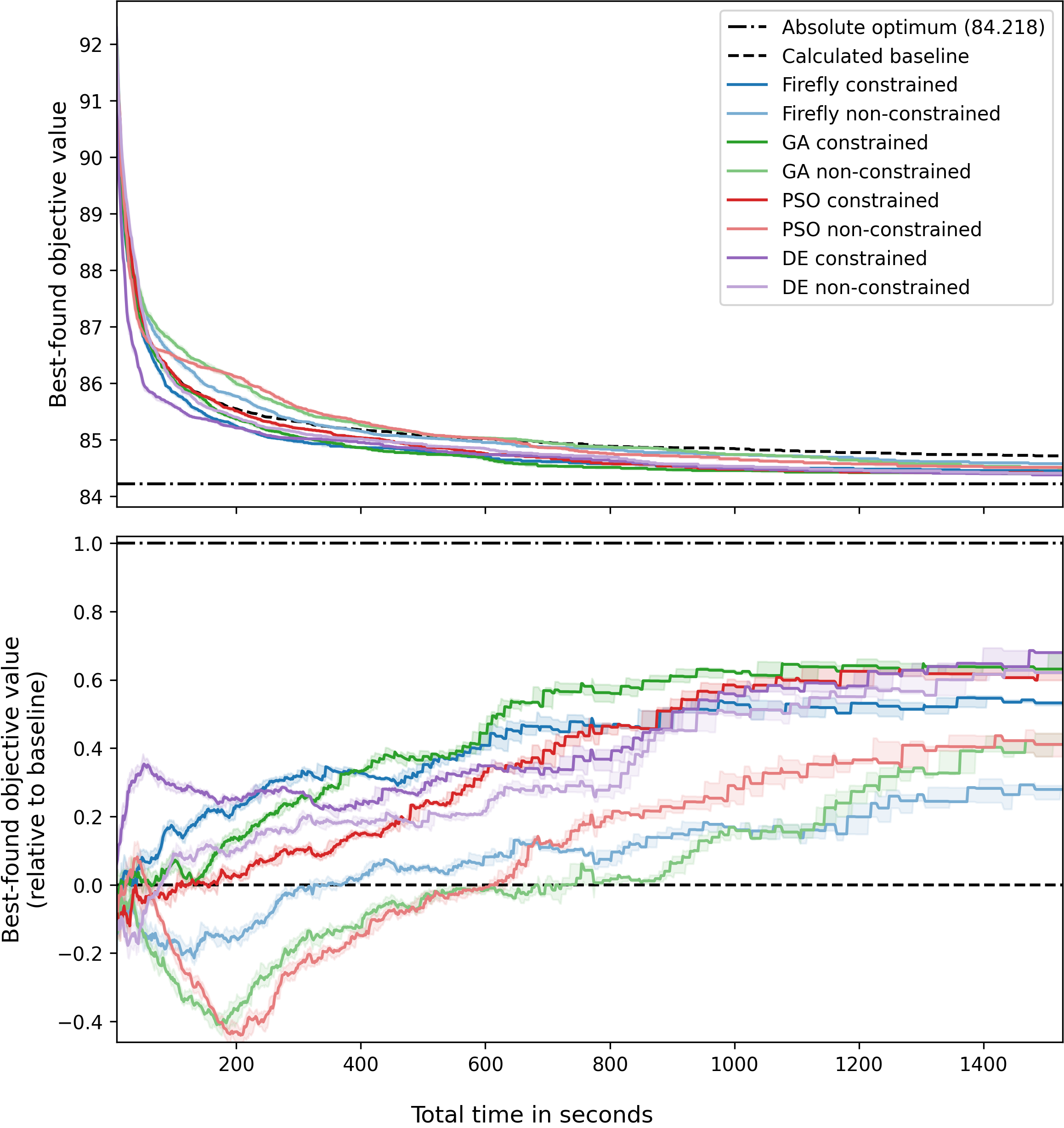}
    \caption{The performance over time of our constraint-aware optimization algorithms implementations and the Kernel Tuner default implementations for the Dedispersion kernel on the Nvidia A6000.}
    \label{fig:evaluation_kt_specific_searchspace}
\end{figure}

Before comparing these optimization algorithms across all 24 search spaces, let us first compare the performance of the constrained-optimized versions of each optimization algorithm to their non-optimized counterpart on a single search space. 
\Cref{fig:evaluation_kt_specific_searchspace} compares this over time in two plots for the Dedispersion kernel on the Nvidia A6000. 
The top plot shows the absolute best-found lowest runtime on this search space for each of the algorithms, up to the absolute optimum of 84.218 milliseconds. 
The bottom plot shows the same data, but relative to the baseline (fixed to 0.0) and the optimum at 1.0, making it easier to distinguish performance differences among the optimization algorithms. 
As mentioned in \cref{subsec:evaluation_experimental_setup}, like all other experiments in this evaluation, the budget is set to the time it takes the calculated random search baseline to reach a configuration that has a performance of at least 95\% of the distance between the search space median and optimum. 
In the bottom plot of \cref{fig:evaluation_kt_specific_searchspace}, we can see that most algorithms start with negative scores, meaning that at the start, the algorithms perform worse than the calculated baseline. 
It can be seen in \cref{fig:evaluation_kt_specific_searchspace} that each constraint-aware optimization algorithm outperforms its non-optimized counterpart by a significant margin. 
It is particularly noteworthy that for PSO, Firefly, and GA, the performance of the non-constrained variants approaches that of the random search baseline for the first half of the tuning time. 

\begin{figure}[tb]
    \centering
    \includegraphics[width=\linewidth]{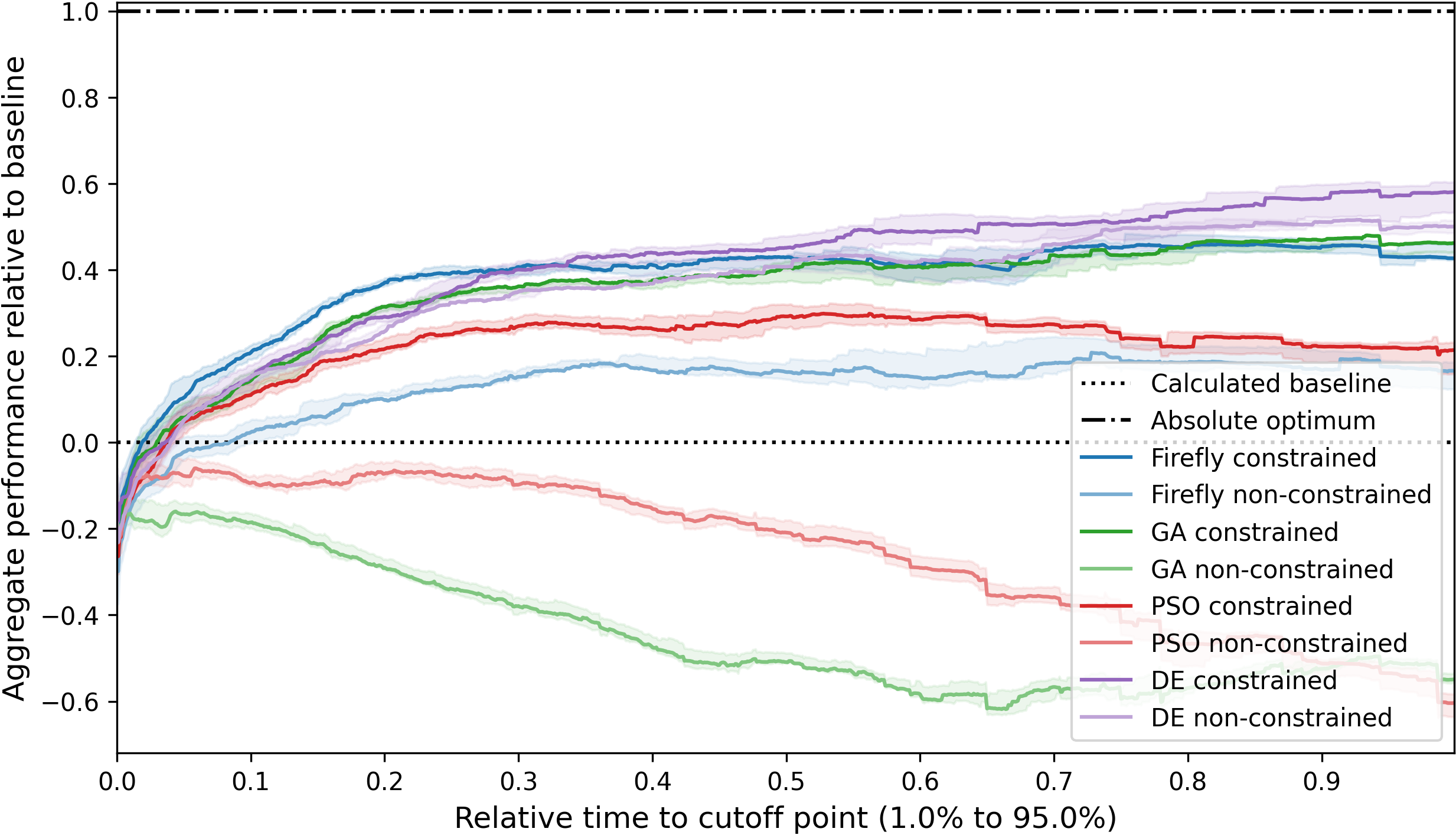}
    \caption{
    The aggregate performance over time of our constraint-aware optimization algorithms implementations and the Kernel Tuner default implementations.}
    \label{fig:evaluation_kt_aggregate_performance}
\end{figure}

To evaluate the overall performance, \cref{fig:evaluation_kt_aggregate_performance} compares the performance of the constrained-optimized versions of each optimization algorithm to their non-optimized counterpart over time across all 24 search spaces.
While Differential Evolution (\textit{DE}) has a minor gain with constraint awareness, the Genetic Algorithm (\textit{GA}), Firefly, and PSO constraint-aware algorithms outperform their counterparts by a wide margin. 
Quantifying this difference with the \textit{performance score} (an optimization algorithm's performance over the passed time relative to the calculated baseline), making Genetic Algorithm constraint-aware improved the score by 0.801, PSO by 0.478, DE by 0.047, and Firefly by 0.245, for an average improvement of 0.393. This can be interpreted as these algorithms finding the same configuration in $\sim39\%$ less time, finding $\sim39\%$ better performing configurations in the same time, or a combination.

\begin{figure}[tbp]
\centering
\subfloat[Firefly non-constrained\label{fig:results_heatmap_firefly_non_constrained}]{%
  \includegraphics[width=0.499\linewidth]{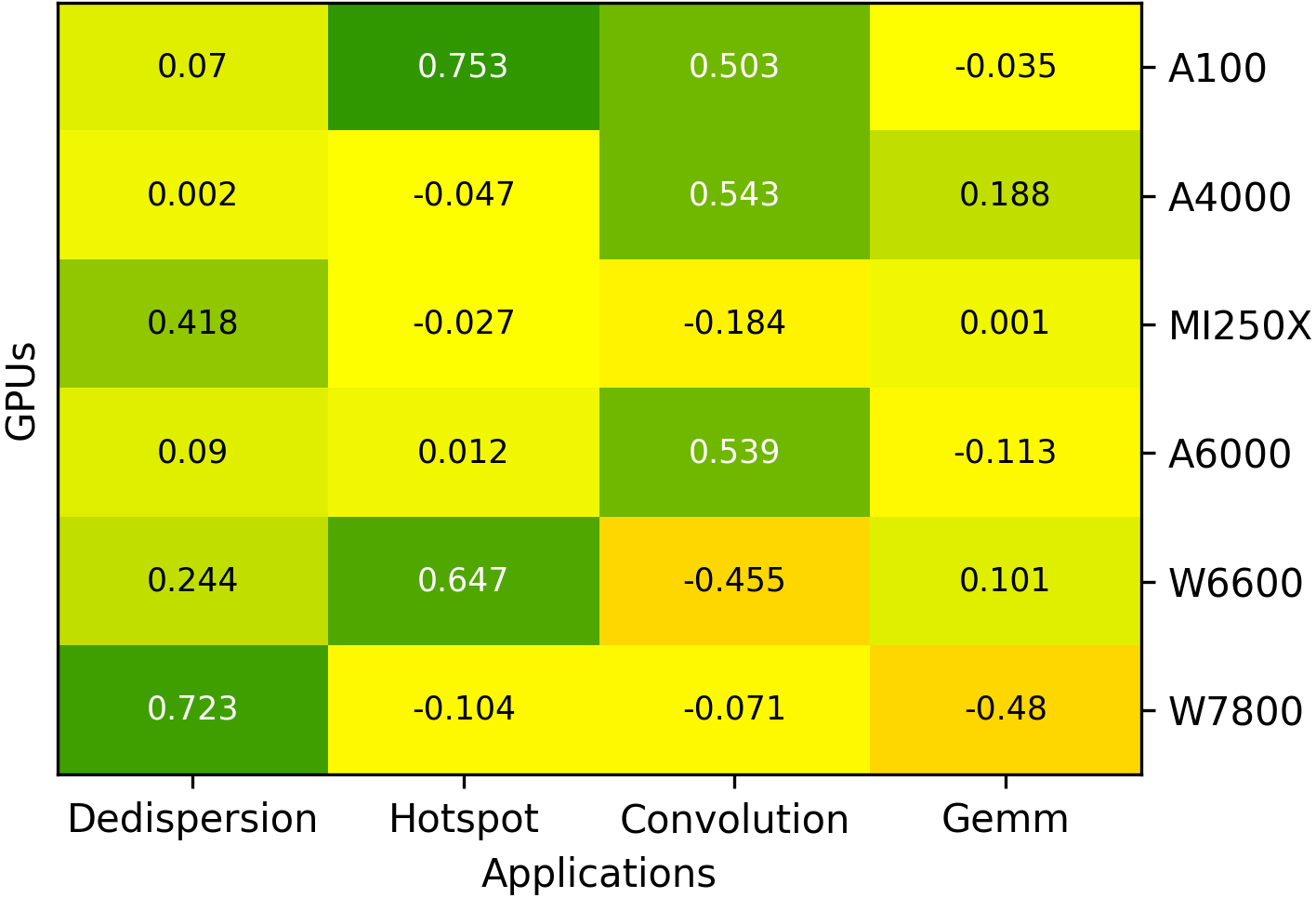}%
}\hfil
\subfloat[Firefly constrained\label{fig:results_heatmap_firefly_constrained}]{%
  \includegraphics[width=0.499\linewidth]{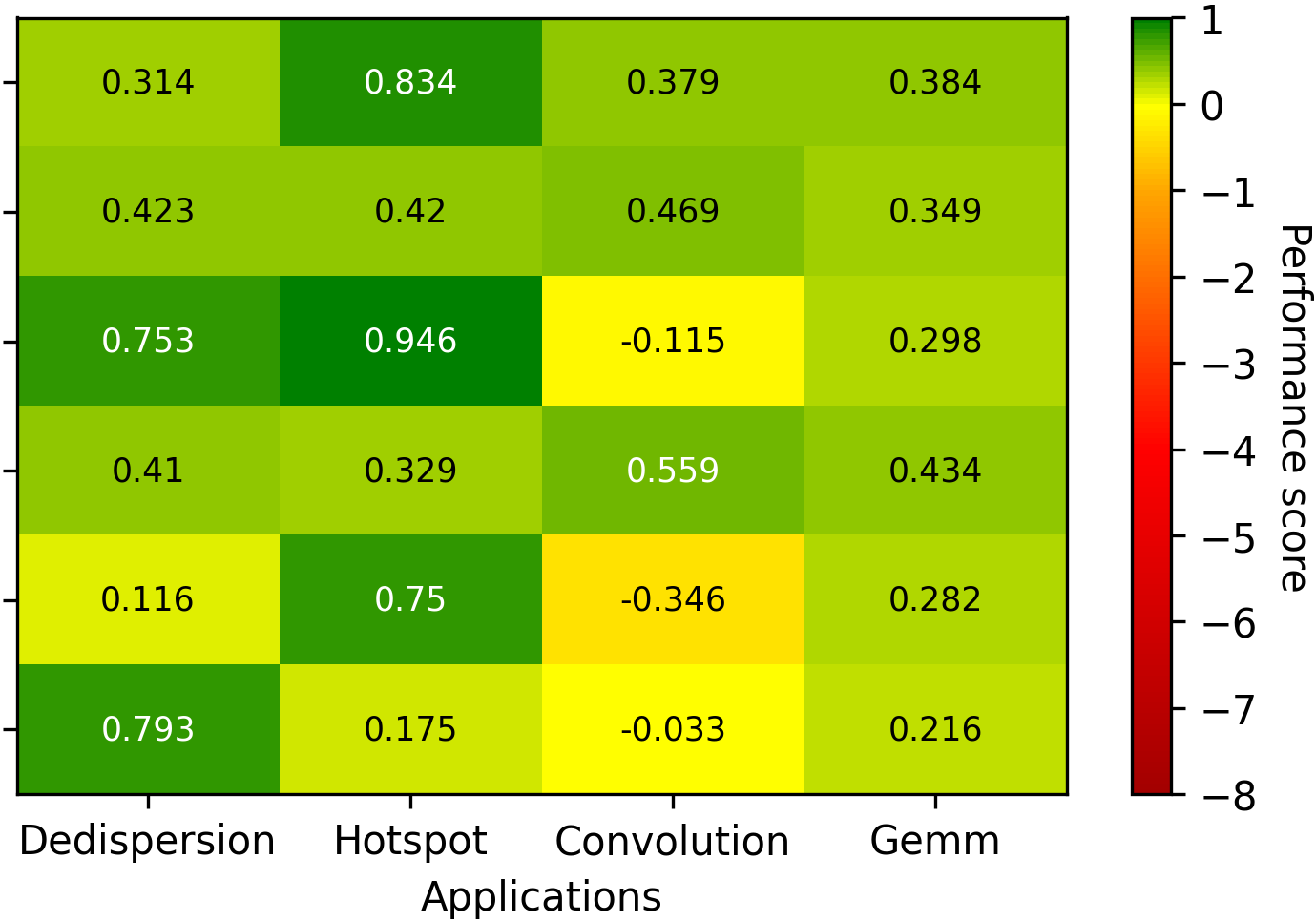}%
} \\
\vspace{-0.3cm}
\subfloat[GA non-constrained\label{fig:results_heatmap_ga_non_constrained}]{%
  \includegraphics[width=0.499\linewidth]{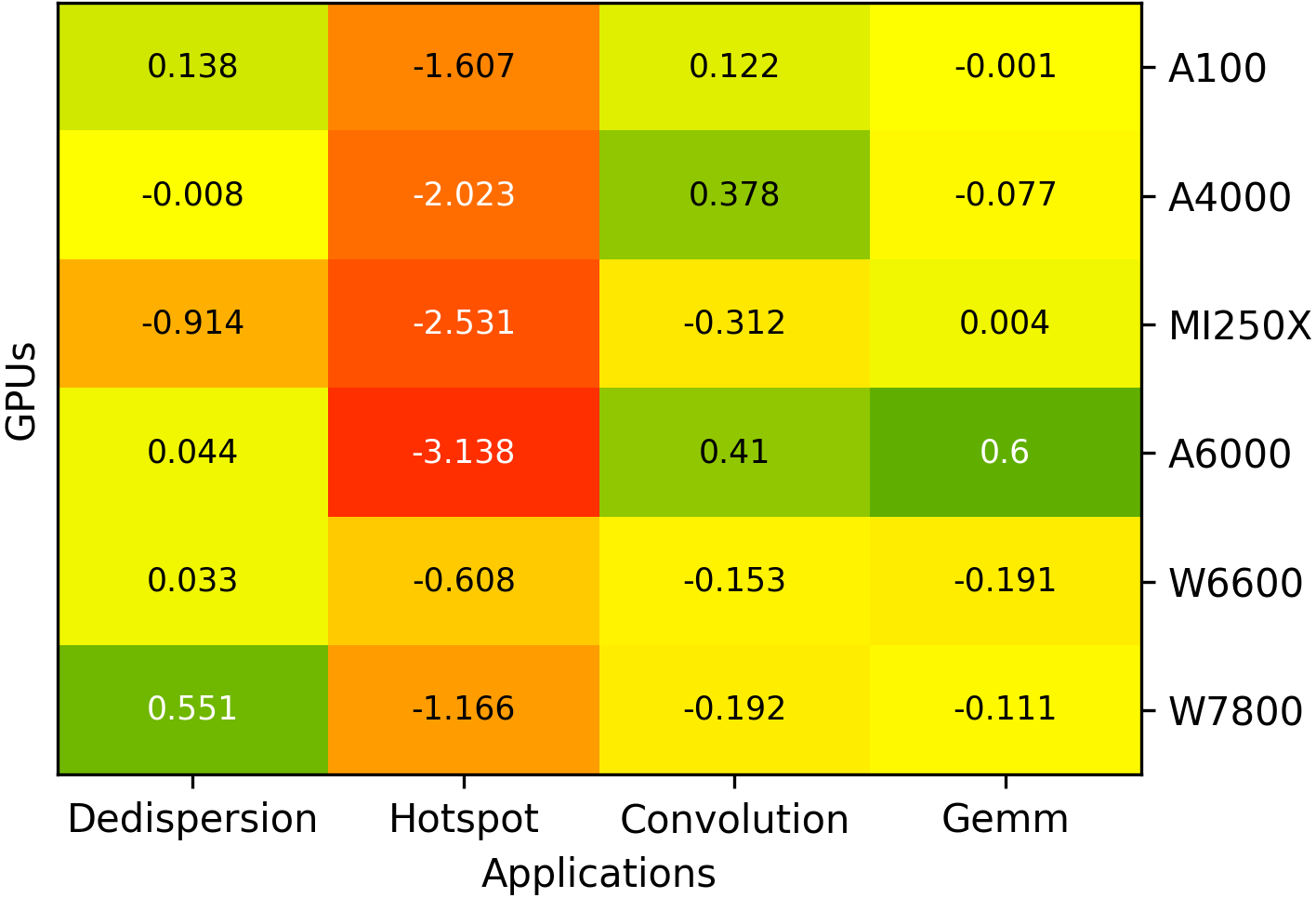}%
}\hfil
\subfloat[GA constrained\label{fig:results_heatmap_ga_constrained}]{%
  \includegraphics[width=0.499\linewidth]{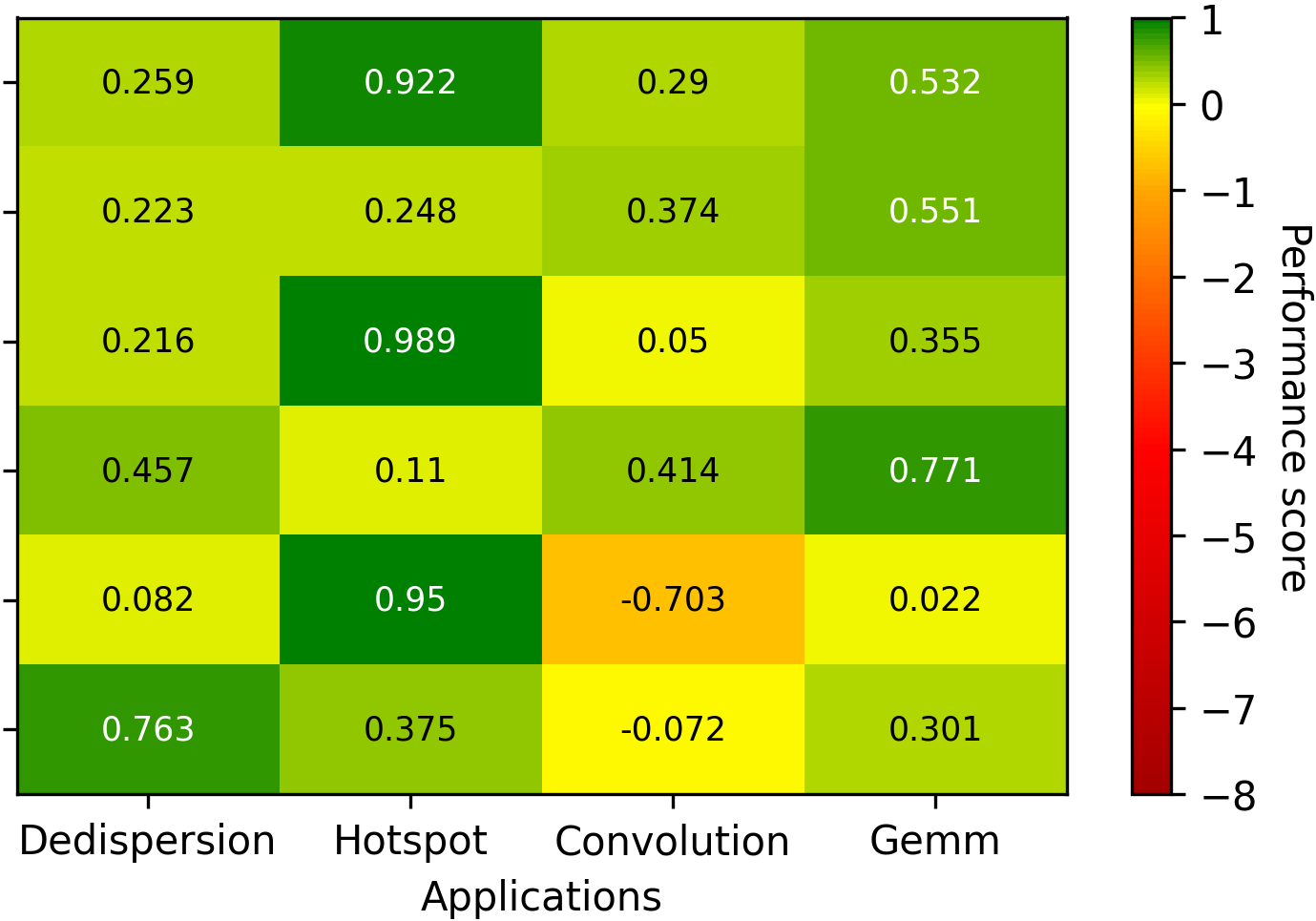}%
} \\
\vspace{-0.3cm}
\subfloat[PSO non-constrained\label{fig:results_heatmap_pso_non_constrained}]{%
  \includegraphics[width=0.499\linewidth]{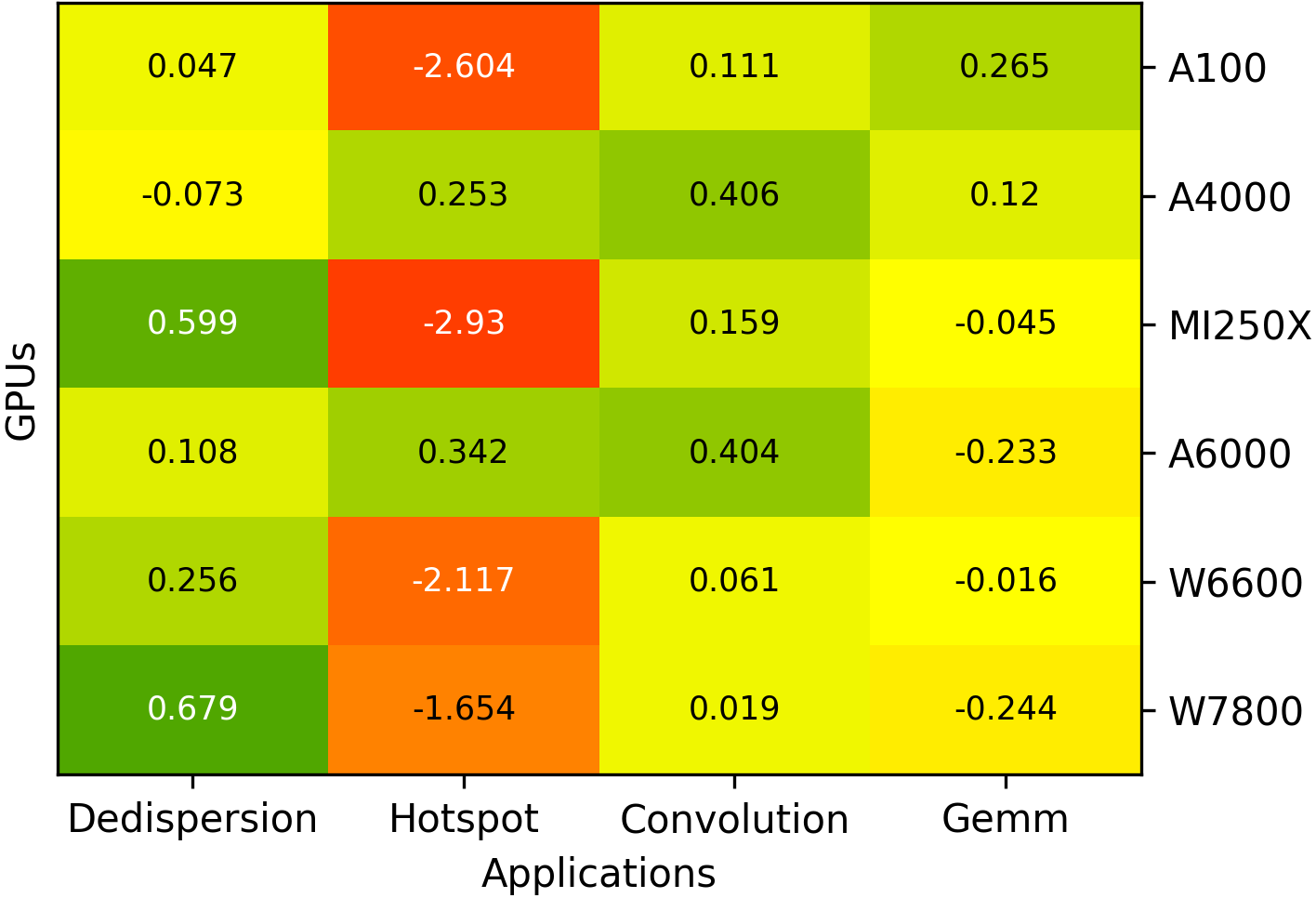}%
}\hfil
\subfloat[PSO constrained\label{fig:results_heatmap_pso_constrained}]{%
  \includegraphics[width=0.499\linewidth]{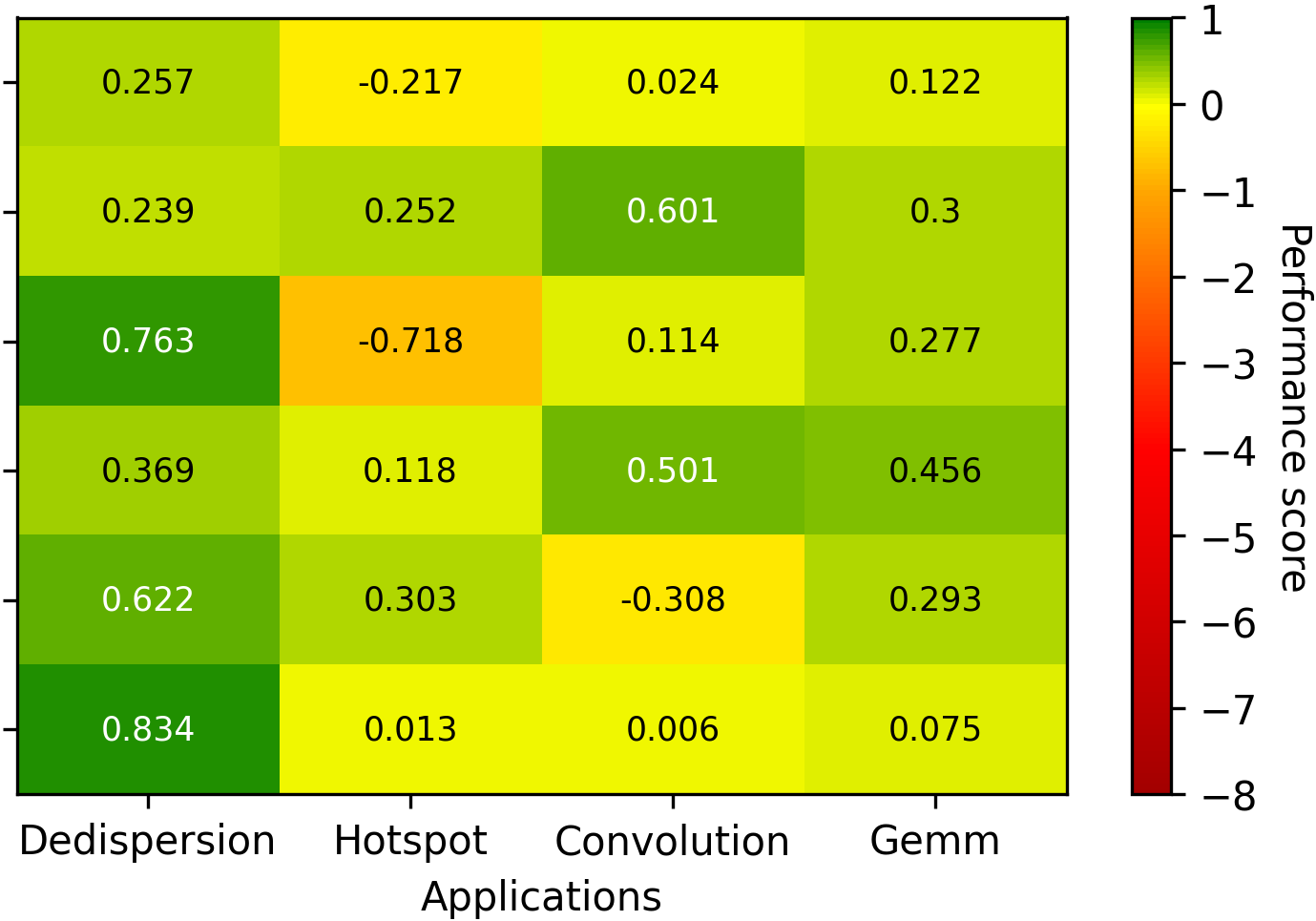}%
} \\
\vspace{-0.3cm}
\subfloat[DE non-constrained\label{fig:results_heatmap_de_non_constrained}]{%
  \includegraphics[width=0.499\linewidth]{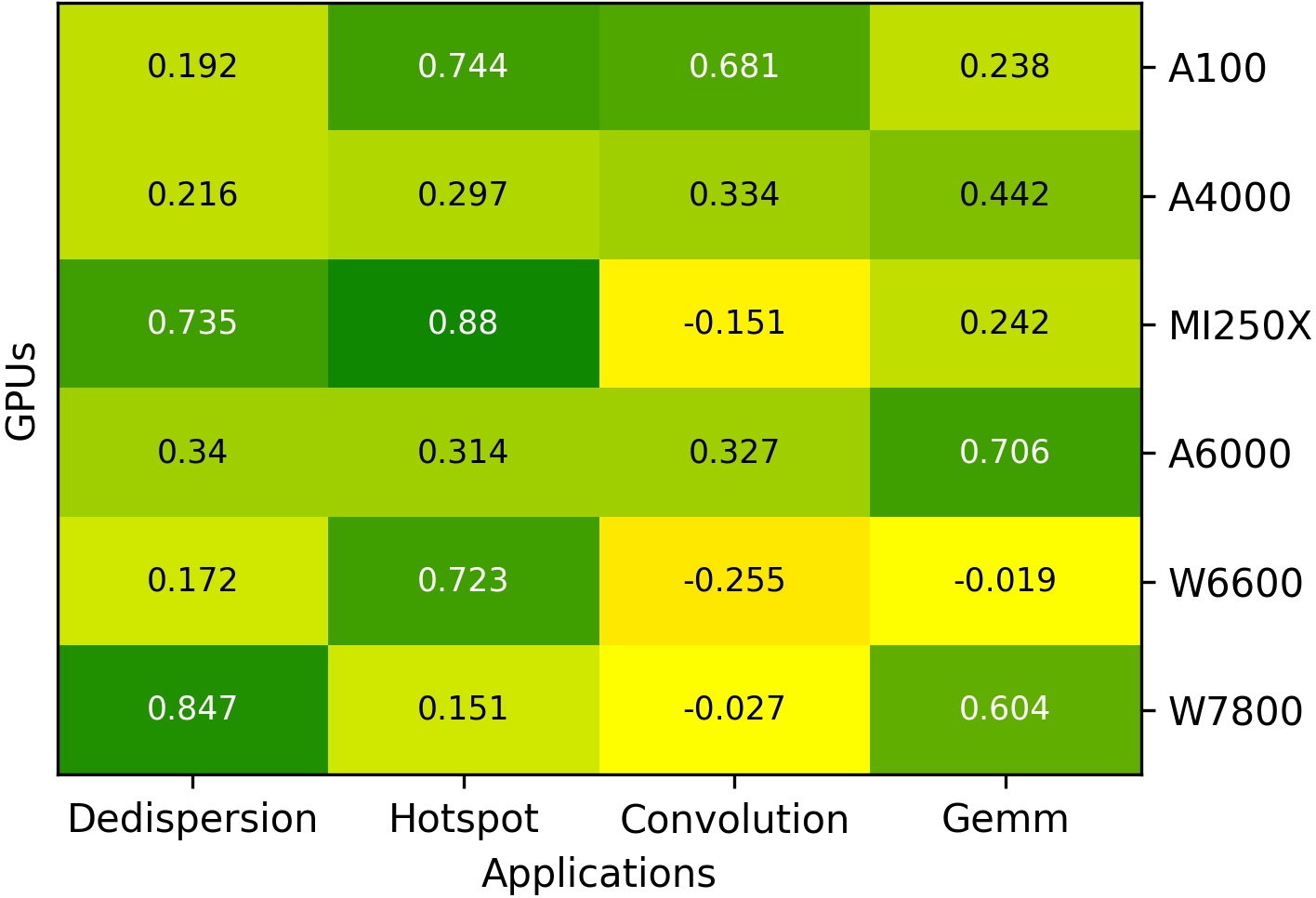}%
}\hfil
\subfloat[DE constrained\label{fig:results_heatmap_de_constrained}]{%
  \includegraphics[width=0.499\linewidth]{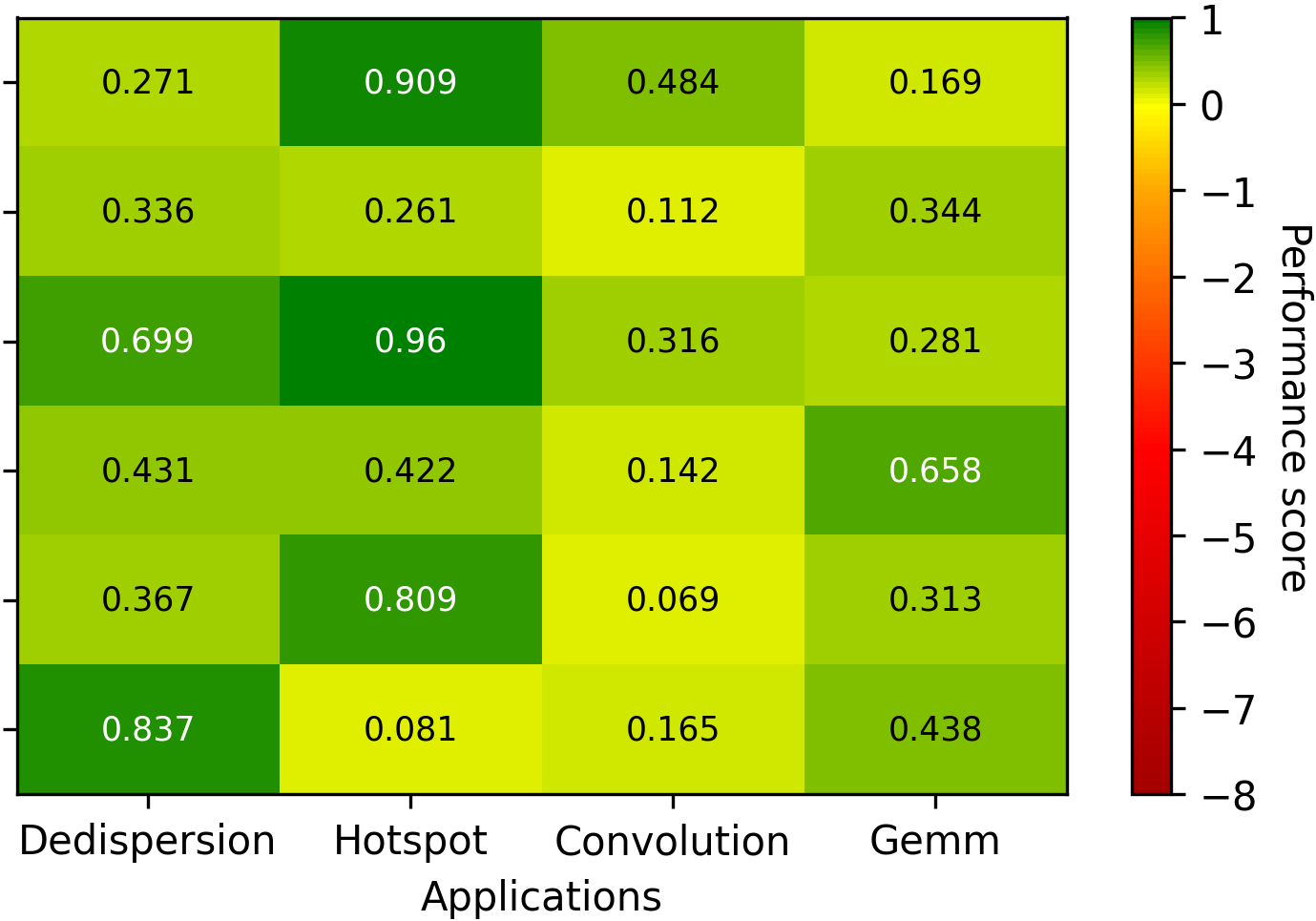}%
}
\caption{Impact of constraint-awareness on optimization algorithm performance per search space.}
\label{fig:results_heatmap_per_searchspace}
\end{figure}

Finally, we can compare the performance score per search space between our constraint-aware and the original versions for specific optimization algorithms to validate our findings, shown in \cref{fig:results_heatmap_per_searchspace}. 
As would be expected when the performance improvement is due to constraint-awareness, the performance difference is greatest between the sparsest search spaces, especially \textit{hotspot} and to a lesser extent \textit{GEMM}, as per \cref{tab:searchspaces_real_world_overview}. This is particularly noticeable for GA and PSO, where the performance difference between both versions was also the largest in \cref{fig:evaluation_kt_aggregate_performance}.

\subsection{Comparison with State of the Art} \label{subsec:evaluation_pyatf}

\begin{figure}[tb]
    \centering
    \includegraphics[width=\linewidth]{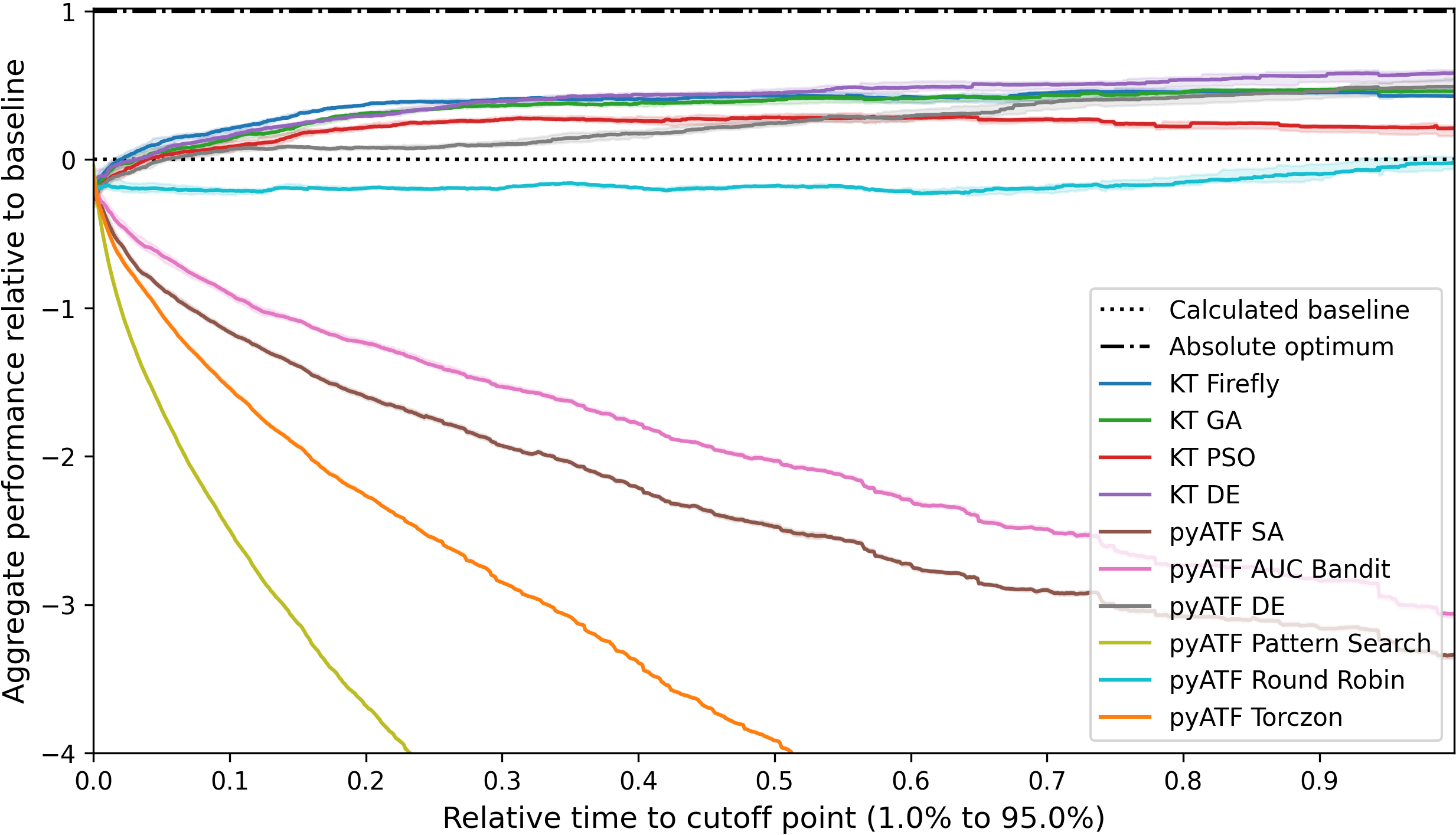}
    \caption{The aggregate performance of various pyATF and our constraint-aware optimization algorithms across all 24 search spaces. The plot has been cut off at -4 to improve legibility, \textit{pyATF Torczon} and \textit{pyATF AUC Bandit} continue to -6 and -10 respectively.}
    \label{fig:evaluation_pyatf_aggregate_performance}
\end{figure}

In this subsection, we evaluate the performance of our constraint-aware optimization algorithms against pyATF~\cite{pyATF}, a state-of-the-art framework for constraint-based auto-tuning. 
As both our methods and pyATF are implemented in Python, we can do a pure substitution of solely the optimization algorithms for a fair comparison. 
As it is known that the chain-of-trees search space generation can be expensive~\cite{searchspace_construction_paper}, we have implemented a retrieval mechanism to prevent this from influencing the results. 
In addition, pyATF does not have a memoization mechanism for retrieving previously evaluated configurations as Kernel Tuner does. To ensure a fair comparison of only the optimization algorithms in both frameworks, we have used Kernel Tuner's cost function for pyATF's methods as well, ensuring that the lack of a memoization mechanism does not put pyATF's methods at a disadvantage. Moreover, this approach also ensures all optimization algorithms in our comparison use the same backends, compilers, and time measurement method.
We compare against all optimization algorithms in pyATF with default hyperparameters; of these, \textit{simulated annealing}, \textit{differential evolution}, \textit{Pattern Search}, and \textit{Torczon} are stand-alone algorithms, while \textit{AUC Bandit} and \textit{Round Robin} are ensemble-based approaches re-using the stand-alone algorithms. 

We show the results of this comparison in \cref{fig:evaluation_pyatf_aggregate_performance}, where it can be seen that of the six pyATF algorithms, only \textit{differential evolution} (abbreviated to \textit{DE}) performs on par with our constraint-aware optimization algorithms, while especially \textit{AUC Bandit}, \textit{Simulated Annealing} (\textit{SA}), \textit{Torzon} and \textit{Pattern Search} perform notably worse than even the non-constrained Kernel Tuner optimization algorithms shown in \cref{fig:evaluation_kt_aggregate_performance}. 
Quantifying this with the performance score, the average score of our optimization algorithms is $0.342$, as opposed to $-2.361$ for pyATF. 

\begin{figure}[tbp]
\centering
\subfloat[pyATF Differential Evolution\label{fig:results_heatmap_pyatf_de}]{%
  \includegraphics[width=0.499\linewidth]{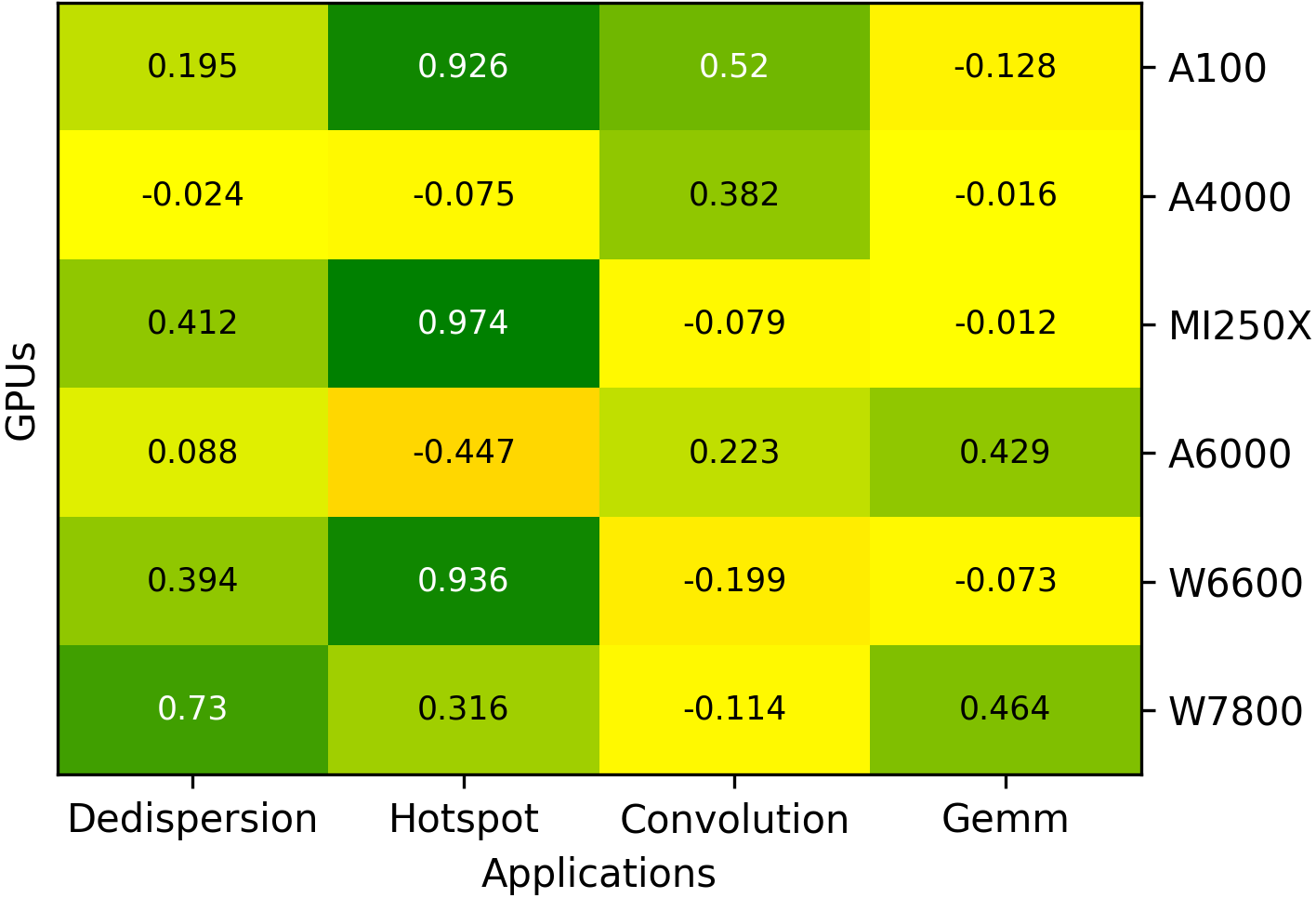}%
}\hfil
\subfloat[Our constraint-aware  DE\label{fig:results_heatmap_kt_de_constrained}]{%
  \includegraphics[width=0.499\linewidth]{figures/evaluation/diff_evo_constrained_heatmap_applications_gpus_colorbar.png}%
} \\
\vspace{-0.3cm}
\subfloat[pyATF Round Robin\label{fig:results_heatmap_pyatf_round_robin}]{%
  \includegraphics[width=0.499\linewidth]{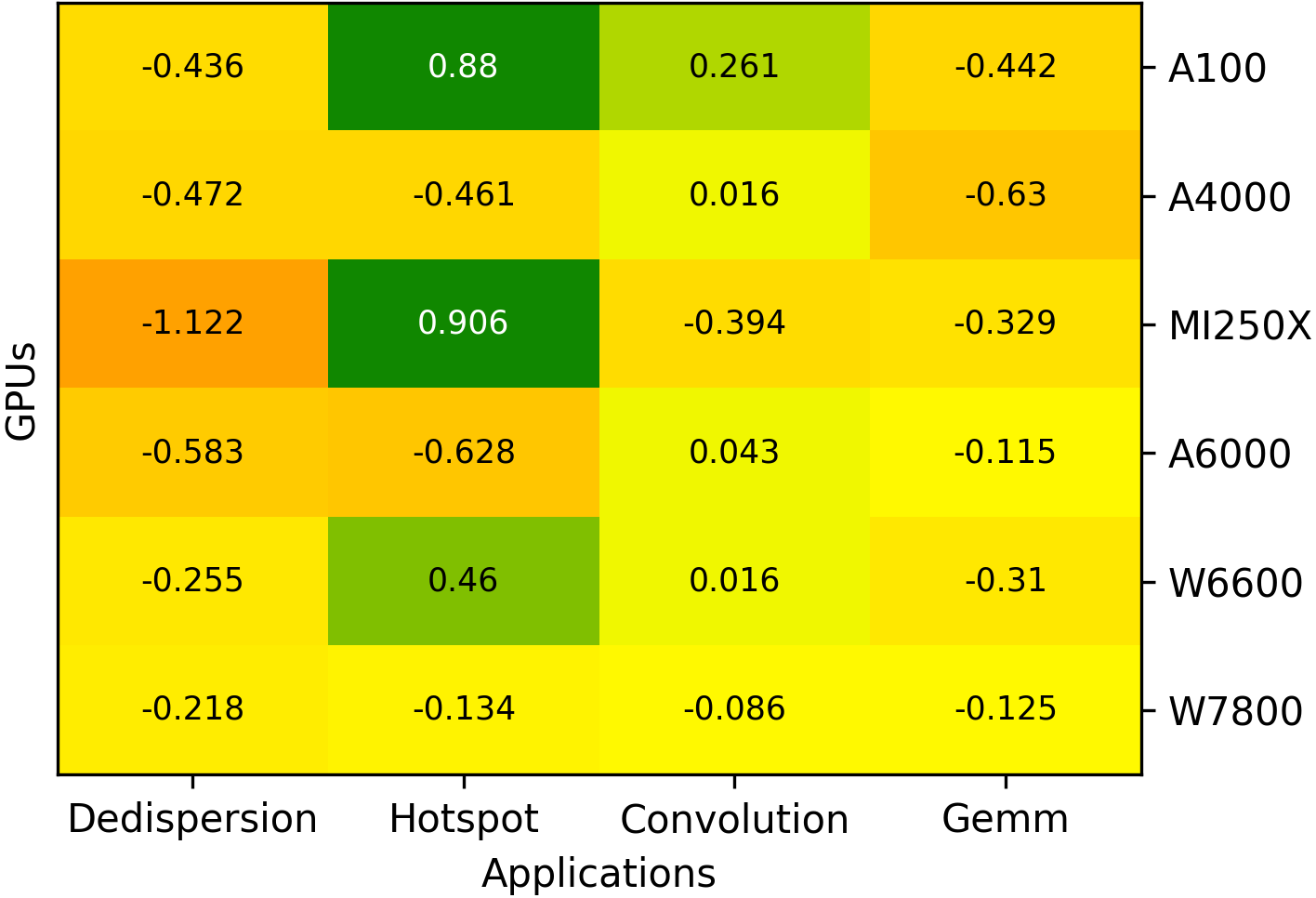}%
}\hfil
\subfloat[Our constraint-aware Firefly\label{fig:results_heatmap_kt_firefly_constrained}]{%
  \includegraphics[width=0.499\linewidth]{figures/evaluation/firefly_constrained_heatmap_applications_gpus_colorbar.png}%
} \\
\caption{Performance difference per search space for \textit{Differential Evolution} and the next best-performing algorithms between pyATF and the constraint-aware implementations presented in this paper.}
\label{fig:results_heatmap_per_searchspace_pyatf_vs_kt}
\end{figure}

To validate our findings, we can compare the performance score per search space between the pyATF and the constraint-aware optimization algorithms we introduce in this paper. 
This is shown in \cref{fig:results_heatmap_per_searchspace_pyatf_vs_kt} for \textit{Differential Evolution}, which is present in both frameworks, as well as for the next best-performing algorithms. 
In the latter case, our constraint-aware Firefly (\cref{fig:results_heatmap_kt_firefly_constrained}) outperforms pyATF Round Robin (\cref{fig:results_heatmap_pyatf_round_robin}) on all but two of the 24 search spaces. 
The pyATF implementation of \textit{Differential Evolution} (\cref{fig:results_heatmap_pyatf_de}) is outperformed by our DE implementation (\cref{fig:results_heatmap_pyatf_de}) on the majority of search spaces as well. %

\section{Conclusion}\label{sec:conclusion}

Automatic performance tuning is essential in high-performance computing for achieving optimal application performance across diverse hardware platforms. However, the presence of complex parameter interdependencies and hardware constraints introduces a significant challenge in optimization. In this work, we addressed this challenge by integrating constraint-handling strategies into four widely used evolutionary optimization algorithms, Differential Evolution, Particle Swarm Optimization, Firefly, and Genetic Algorithm, within a generic auto-tuning framework.

Our results demonstrate that equipping these algorithms with constraint-awareness leads to substantial performance improvements ($\sim39\%$) compared to their non-constraint-aware counterparts, correlated with search space sparsity. We observed faster convergence, more efficient exploration of the feasible search space, and higher-quality solutions across a broad range of auto-tuning search spaces. Furthermore, we showed that our methods outperform the methods in the state-of-the-art pyATF framework, underscoring the practical value of our approach.
The constrained optimization algorithms presented in this paper have been made available as open-source contributions to the open-source Kernel Tuner framework, lowering the barrier to adoption and enabling reproducibility.

\ifdoubleblind
\else
\section*{Acknowledgment}
\footnotesize{
The CORTEX project has received funding from the Dutch Research Council (NWO) in the framework of the NWA-ORC Call (file NWA.1160.18.316).
The ESiWACE3 project has received funding from the European High Performance Computing Joint Undertaking (JU) under grant agreement No 101093054.
}
\fi

\bibliographystyle{IEEEtran.bst}
\bibliography{library}

\end{document}